%% file: 0-main.tex
\documentclass[lettersize,journal]{IEEEtran}
\usepackage{xr-hyper}

\input{premble}

\begin{document}

\title{Personalized Federated Social Bot Detection via Cooperative Reinforced Contrastive Adversarial Distillation}

\author{Yingguang Yang,
        Hao Liu,
        Xin Zhang,
        Yunhui Liu,
        Yutong Xia,
        Qi Wu,
        Hao Peng,
        Taoran Liang,
        Bin Chong,
        Tieke He,
        Philip S. Yu,~\IEEEmembership{Life Fellow,~IEEE},
\IEEEcompsocitemizethanks{
\IEEEcompsocthanksitem Yingguang Yang, Hao Liu, Xin Zhang and Qi Wu are with the School of Cyber science at University of Science and Technology of China, Hefei 230026, China, Guangxi Key Lab of Multi-source Information Mining \& Security, Guangxi Normal University, Guilin 541004, China. E-mail: \{dao, rcdchao, darcy , qiwu4512\}@mail.ustc.edu.cn;
\IEEEcompsocthanksitem Yutong Xia is with the Institute of Data Science at the National University of Singapore, 119077, Singapore. E-mail: yutong.xia@u.nus.edu;
\IEEEcompsocthanksitem Hao Peng is with the State Key Laboratory of Software Development Environment, Beihang University, Beijing 1000191, China, Guizhou Provincial Artificial Intelligence Laboratory, Guiyang 550000, China. E-mail: \{penghao\}@buaa.edu.cn;
\IEEEcompsocthanksitem Taoran Liang is with the in Systems Science from Beijing University of Posts and Telecommunications, Beijing 100876, China. E-mail: taorann@bupt.edu.cn.
\IEEEcompsocthanksitem Bin Chong is with the National Engineering Laboratory for Big Data Analysis and Applications, Peking University, Beijing, 100871, P.R. China. E-mail: chongbin@pku.edu.cn;
\IEEEcompsocthanksitem Yunhui Liu and Tieke He are with the State Key Laboratory for Novel Software Technology, Nanjing University, Nanjing 210023, China. E-mail: \{lyhcloudy1225, hetieke\}@gmail.com;
\IEEEcompsocthanksitem Philip S. Yu is with the Department of Computer Science, University of Illinois Chicago, Chicago, IL 60607, USA. E-mail: psyu@uic.edu.
}
\thanks{Manuscript received September 2025. (Corresponding authors: Hao Peng, Bin Chong and Tieke He.)}
}

\markboth{IEEE Transactions on Knowledge and Data Engineering}%
{Yang \MakeLowercase{\textit{et al.}}: Personalized Federated Social Bot Detection via Reinforced Cooperative Aggregation Unite with Contrastive Adversarial Distillation}

  \IEEEtitleabstractindextext{%
\begin{abstract}
Social bot detection is critical to the stability and security of online social platforms.
However, current state-of-the-art bot detection models are largely developed in isolation, overlooking the benefits of leveraging shared detection patterns across platforms to improve performance and promptly identify emerging bot variants.
The heterogeneity of data distributions across platforms further complicates the design of an effective cross-platform detection framework.
To address these challenges, we propose \FedRio (Personalized \underline{Fed}erated Social Bot Detection with Cooperative \underline{R}e\underline{i}nf\underline{o}rced Contrastive Adversarial Distillation).
We first introduce an adaptive message-passing module as the graph neural network backbone for each client.
To facilitate efficient knowledge sharing of global data distributions, we design a federated knowledge extraction mechanism based on generative adversarial networks.
Additionally, we employ a multi-stage adversarial contrastive learning strategy to reduce the divergence between local and global models.
Finally, we adopt neuron-level server-side parameter aggregation and a client-side controller, trained by reinforcement learning, that sets in each round how much of the global model a client absorbs.
Extensive experiments on two real-world social bot detection benchmarks, run over twenty seeds with validation-selected checkpoints and paired significance tests, demonstrate that \FedRio outperforms state-of-the-art federated learning baselines in global detection accuracy, communication cost, and feature space consistency, leading the strongest federated baseline by up to $7.4$ macro-F1 points with the margin widening as label skew sharpens, while remaining within $1.6$ accuracy points of the strongest centralized detector although every client sees only its own accounts and the edges among them.
\end{abstract}

}

\maketitle
\IEEEdisplaynontitleabstractindextext
\IEEEpeerreviewmaketitle

\input{1-introduction}
\input{2-literatures}
\input{3-preliminaries}
\input{4-methodology}
\input{5-experiment-setup}

\input{6-experiment}

\section*{Acknowledgments}
This work was supported by the National Natural Science Foundation of China (NSFC) (Grant No.U25B2029, No.62432006 and No.62322202), the Research Fund of Guangxi Key Lab of Multi-source Information Mining \& Security (MIMS23-M-01), the Major Science and Technology SpecialProject of Guizhou Provincial Artificial Intelligence Laboratory (Grant No. QiankehePlatform RSSYS[2025] Major 004).

\bibliographystyle{IEEEtran}
\bibliography{Bibliography}

\begin{IEEEbiography}
[{\includegraphics[width=1in,height=1.25in,clip,keepaspectratio]{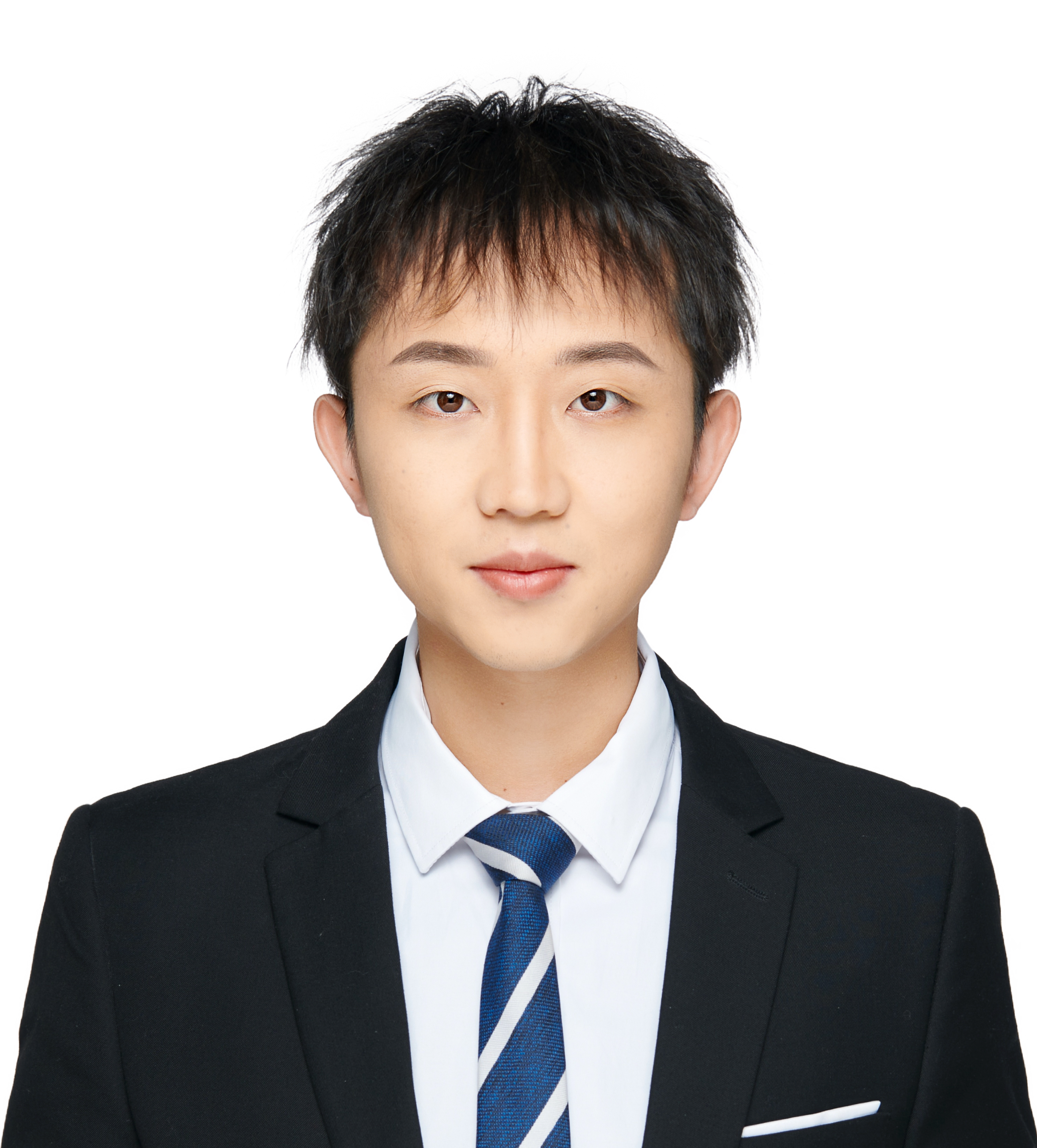}}]
{Yingguang Yang} received the Ph.D. degree in the School of Cyber Science at University of Science and Technology of China.
His research interests include NLP, LLM, data mining, and social bot detection.
\end{IEEEbiography}

\begin{IEEEbiography}
[{\includegraphics[width=1in,height=1.25in,clip,keepaspectratio]{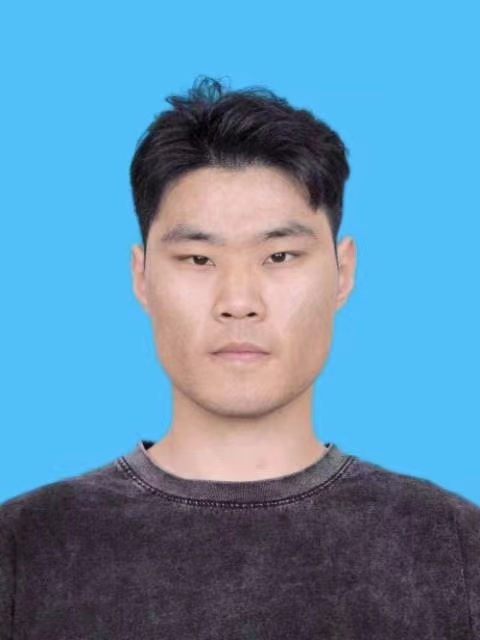}}]
{Hao Liu} is currently a master's student in the School of Cyber Science and Technology at the University of Science and Technology of China. His research interests include fairness, federated learning and graph neural networks.
\end{IEEEbiography}

\begin{IEEEbiography}
[{\includegraphics[width=1in,height=1.25in,clip,keepaspectratio]{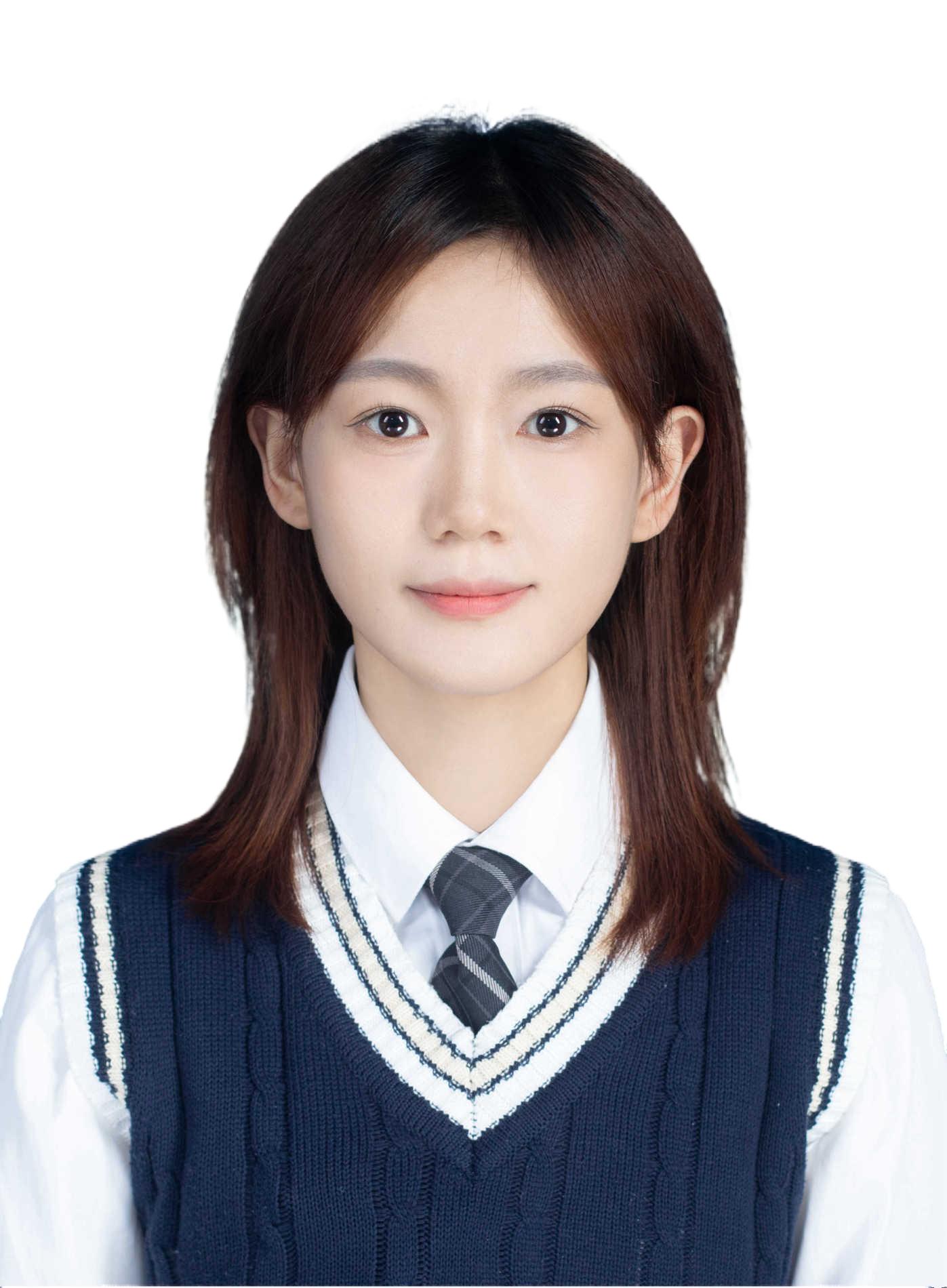}}]
{Xin Zhang} is currently a master's student in the School of Cyber Science and Technology at the University of Science and Technology of China. Her research interests include fairness, federated learning and privacy policy security under the guidance of LLM.
\end{IEEEbiography}

\begin{IEEEbiography}
[{\includegraphics[width=1in,height=1.25in,clip,keepaspectratio]{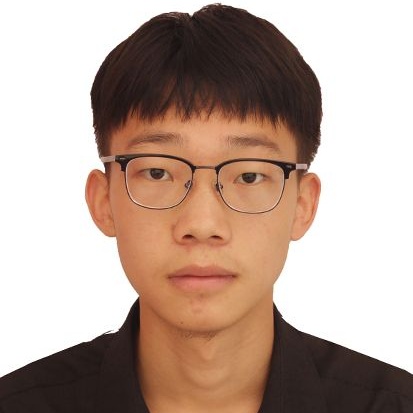}}]
{Yunhui Liu} is working toward the Ph.D. degree with the Software Institute, Nanjing University, China. His research interests include graph machine learning and self-supervised learning.
\end{IEEEbiography}

\begin{IEEEbiography}
[{\includegraphics[width=1in,height=1.25in,clip,keepaspectratio]{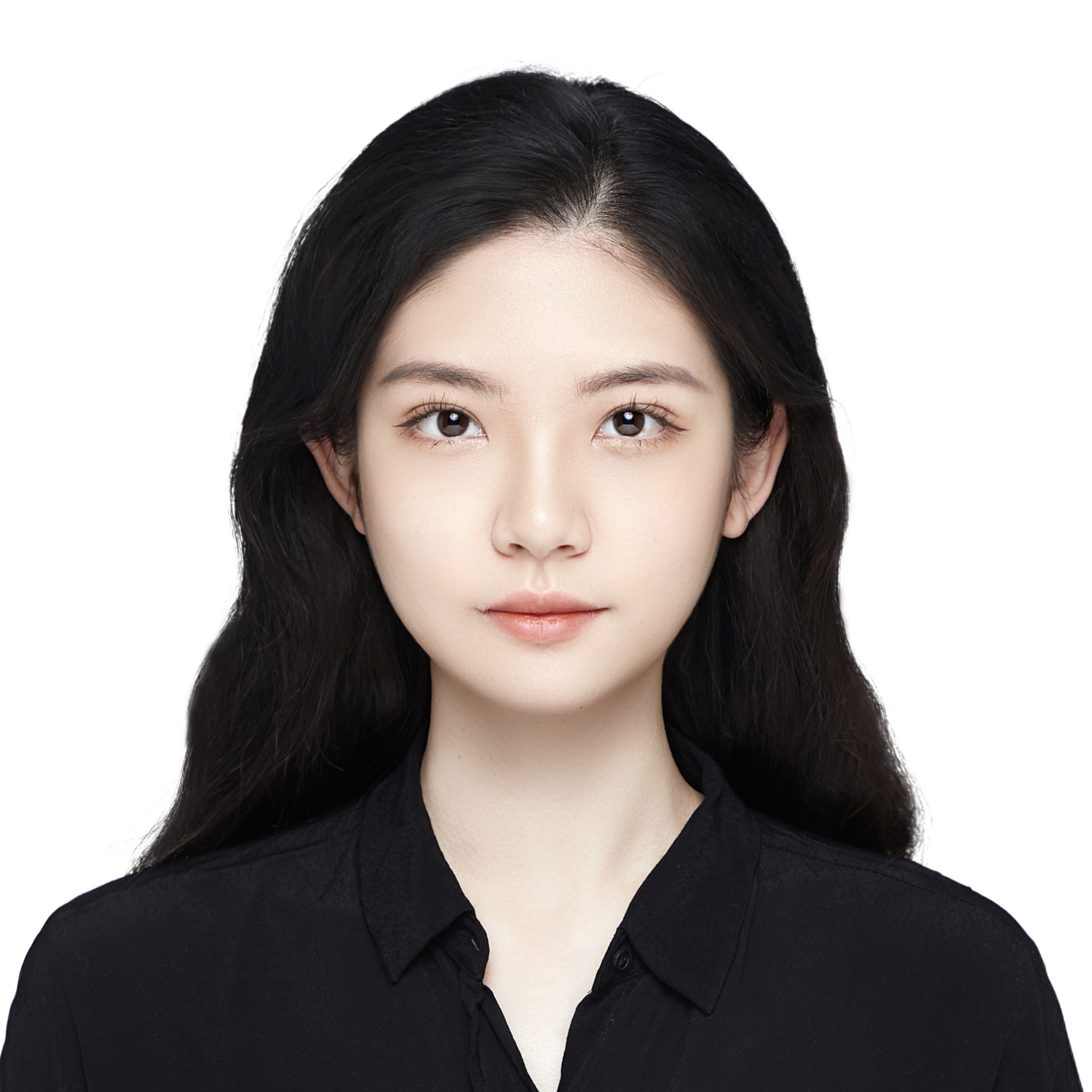}}]
{Yutong Xia}
 is currently a PhD student at the Institute of Data Science, National University of Singapore. She has published several papers in refereed conferences, such as NeurIPS, IJCAI, AAAI, and SIGSPATIAL. Her research interests mainly lie in spatio-temporal data mining and graph neural networks.
\end{IEEEbiography}

\begin{IEEEbiography}
[{\includegraphics[width=1in,height=1.25in,clip,keepaspectratio]{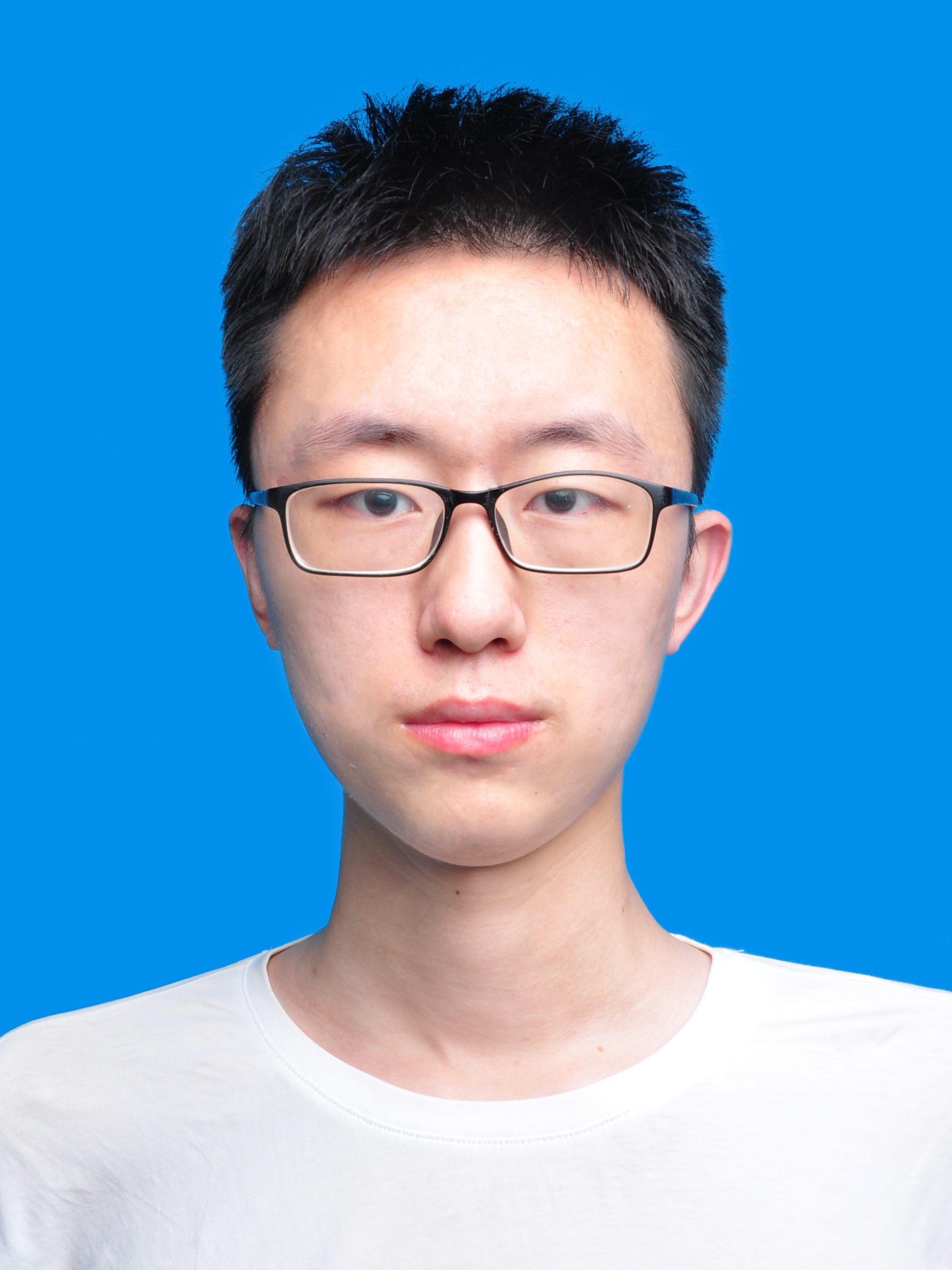}}]
{Qi Wu} is currently a master's student in the School of Cyber Science and Technology at University of Science and Technology of China. His research interests include  multi-view learning, social bot detection and graph neural networks.
\end{IEEEbiography}

\begin{IEEEbiography}
[{\includegraphics[width=1in,height=1.25in,clip,keepaspectratio]{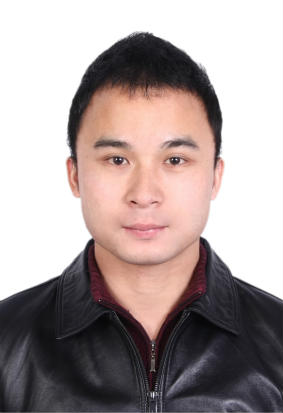}}]
{Hao Peng} is currently a Professor at the School of Cyber Science and Technology in Beihang University.
His current research interests include machine learning, deep learning, and reinforcement learning.
He is the Associate Editor of the International Journal of Machine Learning and Cybernetics (IJMLC).
\end{IEEEbiography}

\begin{IEEEbiography}
[{\includegraphics[width=1in,height=1.25in,clip,keepaspectratio]{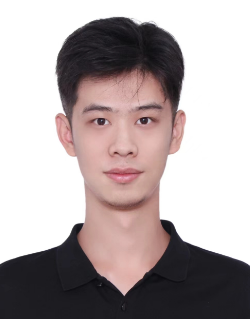}}]
{Taoran Liang} received the M.Sc. degree in Systems Science from Beijing University of Posts and Telecommunications, China. His research interests include reinforcement learning for large language models (LLMs) and graph neural networks (GNNs).
\end{IEEEbiography}

\begin{IEEEbiography}
[{\includegraphics[width=1in,height=1.25in, clip, keepaspectratio]{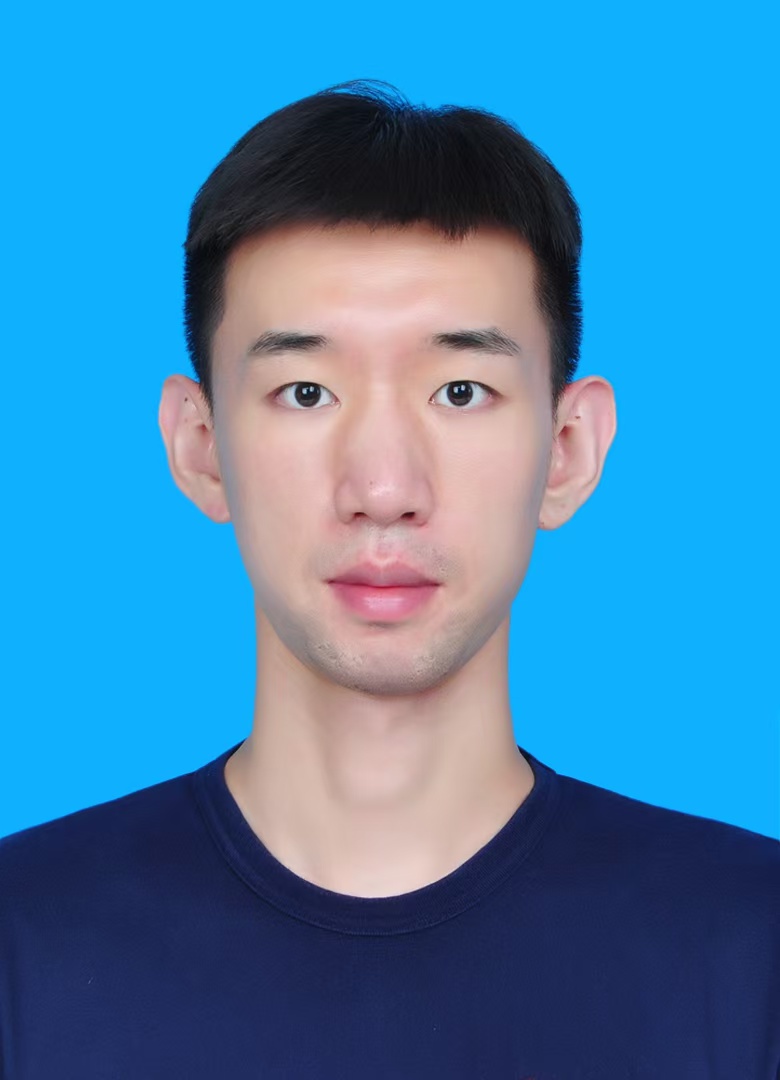}}]
{Bin Chong} received the Ph.D. degree in physical chemistry from the Peking University. He is currently an associate professor in the Peking University. His research interest includes machine learning, molecular modelling and digital transformation.
\end{IEEEbiography}

\begin{IEEEbiography}
[{\includegraphics[width=1in,height=1.25in,clip,keepaspectratio]{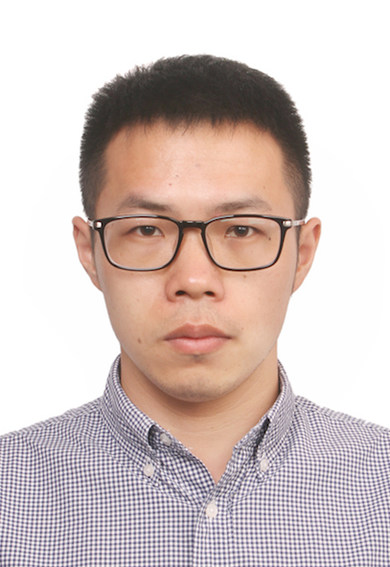}}]
{Tieke He} received the B.E. and Ph.D. degrees in software engineering from the Software Institute, Nanjing University, Jiangsu, China. He is currently an Associate Professor with the Software Institute, Nanjing University. His research interests lie in intelligent software engineering, knowledge graph, and question answering.
\end{IEEEbiography}

\begin{IEEEbiography}
[{\includegraphics[width=1in,height=1.25in,clip,keepaspectratio]{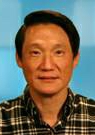}}]
{Philip S. Yu} is a Distinguished Professor and the Wexler Chair in Information Technology at the Department of Computer Science, University of Illinois Chicago.
He is a Fellow of the ACM and IEEE.
Dr. Yu has published more than 1,200 referred conference and journal papers cited more than 190,000 times with an H-index of 195. He has applied for more than 300 patents.
Dr. Yu was the Editor-in-Chief of ACM TKDD (2011-2017) and IEEE TKDE (2001-2004).
\end{IEEEbiography}

\clearpage
\appendices
\setcounter{algocf}{0}\renewcommand{\thealgocf}{S\arabic{algocf}}
\setcounter{table}{0}\renewcommand{\thetable}{S\arabic{table}}
\setcounter{figure}{0}\renewcommand{\thefigure}{S\arabic{figure}}
\section*{APPENDIX}
This appendix collects the pseudocode of the two component procedures, the per-cell
operating points, and the validation-usage controls referred to in the main text.

\subsection*{Complexity and Practicality Analysis}

\vspace*{-2pt}
\subsubsection*{Complexity Analysis}
\label{fedrio:complexity}
The overall complexity of \FedRio consists of the local GNN backbone execution, adversarial knowledge distillation, and the server-side aggregation together with the reinforcement-learned update-rate controller.
For a client graph $\mathcal{G} = (\mathcal{V}, \mathcal{E})$ with $N = |\mathcal{V}|$ nodes and $M = |\mathcal{E}|$ edges, the time complexity of the adaptive message passing backbone $\varepsilon$ is $O(L(N d_1^2 + M d_1))$, where $L$ is the number of layers and $d_1$ is the feature dimension. 
The multi-stage adversarial contrastive knowledge distillation requires $O(E_{local} \cdot N_b \cdot d_2^2)$ time per client, where $E_{local}$ is the number of local epochs, $N_b$ is the batch size, and $d_2$ is the hidden layer size. 
On the server side, one aggregation costs $O(K \cdot |\Theta_{D_1}|)$, where $K$ is the number of clients and $|\Theta_{D_1}|$ is the number of parameters in the shared classifier $D_1$, and the mask optimization repeats this together with a forward and backward pass over $B$ synthetic samples for each of its $S$ steps, giving $O\!\left(S\,(K |\Theta_{D_1}| + B d_2^2)\right)$; note that the masks are defined over $D_1$ only, since the backbone is combined by plain weighted averaging.
The space complexity on the client side is dominated by the storage of the backbone $\varepsilon$, the shared and personalized classifiers $D_1$ and $D_2$, the client generator $G_k$, and the global generator, and is bounded by $O(N d_1 + M + |\Theta_{\varepsilon}| + |\Theta_{D_1}| + |\Theta_{D_2}| + |\Theta_{G_k}| + |\Theta_{G}|)$. The server space complexity is $O(K \cdot |\Theta_{D_1}| + |\Theta_{G}|)$, where $|\Theta_{G}|$ is the parameter size of the global generator. Each client uploads the backbone together with the shared classifier, giving an uplink cost of $O(|\Theta_{\varepsilon}| + |\Theta_{D_1}|)$, while the downlink additionally carries the global generator used for distillation, at $O(|\Theta_{\varepsilon}| + |\Theta_{D_1}| + |\Theta_{G}|)$. The personalized head $D_2$ and the client generator $G_k$ are never transmitted. No raw graph data ever leaves a client, so per-round communication scales with model size rather than with graph size; Section~\ref{exp:setup} instantiates these bounds for the configuration we evaluate.

\vspace*{-2pt}
\subsubsection*{Justification for Framework Complexity}
While \FedRio integrates multiple components, each module addresses a specific challenge in heterogeneous federated bot detection:
\begin{itemize}[leftmargin=*]
    \item \textbf{RL-based Adaptation:} Simpler alternatives face limitations in our setting: (1) per-client validation-based weighting requires each client to hold a representative held-out split, which is problematic under severe label skew ($\alpha \leq 0.1$) where some clients may lack certain classes entirely; (2) gradient-based heuristics (e.g., gradient cosine similarity) provide only first-order local information and cannot capture the cross-round effect of an update-rate decision. Our RL agent observes a state vector formed from client representations, predicted class probabilities and loss values, and learns a non-myopic policy over update rates.
    \item \textbf{Neuron-level Masking:} Standard client-level weighting assumes every coordinate of a client's update is equally trustworthy. Neuron-level masking instead learns where each client's update should be admitted, which is what allows a client holding most of a class to dominate the coordinates that encode it without dominating the rest of the classifier.
\end{itemize}
Both claims are design arguments, and we test them rather than assert them: Section~\ref{sec:module-validation} removes each module in turn, replaces the learned controller by four simpler rules, and reports where the added complexity pays for itself and where it does not.

\begin{algorithm}
\small
\caption{Client Parameter Update Based on Reinforcement Learning}\label{fedrio:algorithm3}
\KwIn{Sample feature vectors \(\{r_t^k\}\), predicted probability distributions \(\{\mathbf{p}_t^k\}\), loss values \(\{\ell_t^k\}\), learning rate \(\eta\), discount factor \(\gamma\), constant \(\xi\), target accuracy \(\omega^{*}\), number of clients \(K\), communication round \(t\)}
\KwOut{Client model parameters at the start of round \( t \), \(\Theta^k_{t,0}\)}
Construct the state vector for the current communication round: \[
s_t = [ r_t^1, \mathbf{p}_t^{1}, \ell_t^{1}, \ldots, r_t^K, \mathbf{p}_t^{K}, \ell_t^{K} ]
\]

Select action \( a_t \) based on state \( s_t \) according to Eq.~\eqref{fedrio:eq:pi}\;


\For{\textbf{each client} \( k=1, \ldots, K \) \textbf{in parallel}}{
    Update client parameters using action \( a^k_t \) according to Eq.~\eqref{fedrio:eq:updateclient}\;

}

Evaluate the accuracy \(\omega_t\) of the aggregated global model on the validation set\;

Calculate the reward \( r_t \) according to Eq.~\eqref{fedrio:eq:reward}\;


Store the transition \((s_t, a_t, r_t, s_{t+1})\) in the replay buffer;

\For{\textbf{each reinforcement learning update step}}{
    Sample a batch of transitions \((s, a, r, s')\) from the replay buffer;

    Compute the target Q-value \( Y_t^{\text{DoubleQ}} \) using Eq.~\eqref{fedrio:eq:ddqn}\;


    Minimize the loss function and update the Q-network according to Eq.~\eqref{fedrio:eq:lossRL}\;


    Update network parameters \(\theta_t\) according to Eq.~\eqref{fedrio:eq:updateqnetwork}\;

}

Every \( M \) update steps, update the target network parameters \(\theta_t'\) with the current online network parameters \(\theta_t\);

\end{algorithm}

\begin{algorithm}
\caption{Server-Side Adaptive Parameter Aggregation}\label{fedrio:algorithm2}
\KwIn{Sample labels $\mathbf{y}$, noise $\mathbf{z}$, global generator $G$, learning rate $\eta$, mask steps $S$, validation split $\mathcal{S}_{val}$, uploaded client parameters $\{\Theta^k_t\}$, previous global parameters $\Theta_{t-1}$}
\KwOut{Updated global model parameters $\Theta_t$}

\For{each mask step $s=1,\ldots,S$}{
    Generate a synthetic batch $\tilde{X}$ using Eq.~\ref{fedrio:eq:genx}\; 
    Compute the loss $\mathcal{L}_{agg}$ on $\tilde{X}$ via Eq.~\ref{fedrio:eq:lagg}\;
    Update the mask logits $m_k$ according to Eq.~\ref{fedrio:eq:laggback}\;
    Score the candidate masks on $\mathcal{S}_{val}$; retain the best-scoring state and stop early if it has not improved for $\max(3, \lfloor S/5 \rfloor)$ steps\;\label{fedrio:alg2:guard}
}

Restore the retained mask state and update the server model parameters $\Theta_t$ by Eq.~\ref{fedrio:eq:aggTheta}\; 

\end{algorithm}

\begin{table}[!ht]
\centering
\caption{\FedRio's operating point per cell, selected on validation accuracy. Hidden width is used for both the encoder and the shared head.}
\label{tab:tuned-configs}
\resizebox{0.48\textwidth}{!}{
\begin{tabular}{ll|cccc}
\toprule
Dataset & & $\alpha=1.0$ & $\alpha=0.5$ & $\alpha=0.1$ & $\alpha=0.05$ \\
\midrule
\multirow{4}{*}{TwiBot-20}
 & learning rate & $5\times10^{-4}$ & $5\times10^{-4}$ & $10^{-3}$ & $10^{-3}$ \\
 & hidden width & 128 & 128 & 128 & 64 \\
 & weight decay & $10^{-4}$ & $10^{-4}$ & $10^{-4}$ & $10^{-4}$ \\
 & proximal $\mu$ & 0.1 & 0.001 & 1.0 & 1.0 \\
\midrule
\multirow{4}{*}{MGTAB}
 & learning rate & $3\times10^{-3}$ & $3\times10^{-3}$ & $3\times10^{-3}$ & $3\times10^{-3}$ \\
 & hidden width & 128 & 128 & 64 & 64 \\
 & weight decay & $10^{-2}$ & $10^{-2}$ & $10^{-4}$ & $10^{-4}$ \\
 & proximal $\mu$ & 0.1 & 0.1 & 1.0 & 1.0 \\
\bottomrule
\end{tabular}}
\vspace{-0.5cm}
\end{table}

\begin{table}[!t]
\centering
\caption{Validation-usage controls (macro-F1 \%, mean$\pm$sd over $20$ seeds).
The label-free variant removes the controller reward and label-based mask stopping.
The split-validation protocol assigns \FedRio's internal signals and model selection
to disjoint halves of the validation split. Both controls independently rerun
training under modified validation protocols.
$^{\dagger}$ marks \FedRio\ significantly above the strongest corresponding baseline
at $p<0.05$ under a one-sided paired $t$-test. The label-free variant is compared with
the strongest baseline in the corresponding main-table cell; split-validation \FedRio\
is compared with the baseline in the last column.}
\label{tab:validation-controls}
\resizebox{\columnwidth}{!}{
\begin{tabular}{lc|ccc}
\toprule
Dataset & $\alpha$
& \shortstack{Label-free\\\FedRio}
& \shortstack{Split-validation\\\FedRio}
& \shortstack{Best split-validation\\baseline} \\
\midrule
\multirow{4}{*}{TwiBot-20}
& 1.0
& 85.4$\pm$0.7
& 85.5$\pm$0.7
& \shortstack{85.5$\pm$0.6 (FedDistill)} \\
& 0.5
& 84.6$\pm$1.5
& 84.4$\pm$1.5
& \shortstack{84.7$\pm$1.2 (FedEnsemble)} \\
& 0.1
& 84.2$\pm$1.4$^{\dagger}$
& 83.7$\pm$1.8$^{\dagger}$
& \shortstack{82.1$\pm$3.3 (FED-PUB)} \\
& 0.05
& 83.0$\pm$2.1$^{\dagger}$
& 82.9$\pm$1.7$^{\dagger}$
& \shortstack{77.6$\pm$3.0 (FedDistill)} \\
\midrule
\multirow{4}{*}{MGTAB}
& 1.0
& 85.6$\pm$1.0$^{\dagger}$
& 85.3$\pm$2.1$^{\dagger}$
& \shortstack{83.8$\pm$2.6 (FedEnsemble)} \\
& 0.5
& 84.2$\pm$1.5$^{\dagger}$
& 83.5$\pm$1.9$^{\dagger}$
& \shortstack{76.5$\pm$4.3 (\FedACK)} \\
& 0.1
& 80.5$\pm$2.8$^{\dagger}$
& 80.3$\pm$4.7$^{\dagger}$
& \shortstack{71.0$\pm$5.8 (\FedACK)} \\
& 0.05
& 78.5$\pm$2.5$^{\dagger}$
& 76.4$\pm$4.7$^{\dagger}$
& \shortstack{68.6$\pm$4.5 (\FedACK)} \\
\bottomrule
\end{tabular}}
\end{table}

\end{document}

%% file: premble.tex
\usepackage{pifont}
\usepackage{latexsym}
\usepackage{balance}
\usepackage{newtxmath}
\usepackage{amsmath}
\usepackage[ruled,linesnumbered]{algorithm2e} %
\usepackage{hyperref}
\usepackage{diagbox}
\usepackage{ntheorem}
\usepackage{pgffor}
\usepackage{tikz}
\usepackage{epstopdf}
\usepackage{mathtools}
\usepackage{subfigure}
\usepackage{tabulary}
\usepackage{color}
\usepackage{flushend}
\usepackage{tcolorbox}
\usepackage[utf8]{inputenc}
\usepackage[english]{babel}
\usepackage{tabularx}
\usepackage{microtype}
\usepackage{textcomp}
\usepackage{booktabs}
\usepackage{colortbl}
\usepackage{soul}
\usepackage[switch]{lineno}
\usepackage{xspace}
\usepackage{algpseudocode}
\colorlet{shadecolor}{yellow}
\usepackage{multirow}
\usepackage{graphicx}
\usepackage{enumitem}
\usepackage{stfloats}
\graphicspath{{./pdf/}{./jpeg/}}
\DeclareGraphicsExtensions{.pdf,.jpeg,.png}
\usepackage{array}
\usepackage{mdwmath}
\usepackage{mdwtab}
\usepackage{eqparbox}
\usepackage{url}

\newcommand{\mypara}[1]{{\noindent\textbf{#1}}}
\newcommand{\FedACK}{\textsc{FedACK}\xspace}

\newcommand{\FedRio}{\textsc{FedRio}\xspace}

\AtBeginDocument{%
  \setlength{\abovedisplayskip}{3pt plus 1pt minus 1pt}%
  \setlength{\belowdisplayskip}{3pt plus 1pt minus 1pt}%
  \setlength{\abovedisplayshortskip}{0pt plus 1pt}%
  \setlength{\belowdisplayshortskip}{2pt plus 1pt minus 1pt}%
}

%% file: 1-introduction.tex
\section{Introduction}
Social bots mimic human behavior across platforms such as Twitter, Facebook, and Instagram~\cite{yardi2010detecting}.
Millions of bots, often controlled by automation programs or platform APIs~\cite{abokhodair2015dissecting}, attempt to infiltrate genuine user communities at scale for malicious purposes, including election manipulation~\cite{deb2019perils,ferrara2020characterizing}, disinformation campaigns~\cite{cresci2020decade}, privacy breaches~\cite{varol2017online}, and the spread of extremist ideologies~\cite{berger2015isis,ferrara2016predicting}.
The detrimental societal impacts and compromised user experience on social platforms highlight the urgent need for effective bot detection mechanisms.

A key challenge in social bot detection is the lack of effective privacy-preserving collaboration across platforms.
In practice, bot networks often operate in a coordinated manner across multiple social media platforms, whereas most existing detection approaches remain platform-specific. Prior studies have explored metadata-derived user features~\cite{d2015real,yang2020scalable}, textual content from social posts~\cite{wei2019twitter,feng2021satar}, and more advanced graph-based models~\cite{zhao2020multi,liu2023botmoe,yang2025robctrl,wu2025certainly}. However, these methods are typically developed and evaluated within a single-platform setting, and their effectiveness is therefore constrained by the quantity, structure, and quality of data available on that platform. Moreover, many recent graph-based detectors~\cite{yang2022rosgas,yang2024sebot,he2024dynamicity} take a single platform-wide graph as given. Federated deployment removes that premise without removing the graph: each participant sees only the edges among its own users, and every edge that crosses between participants disappears from view entirely. What federation leaves behind is therefore not a smaller version of the same problem but a different one, in which message passing must work on fragments. These limitations motivate the need for collaborative bot detection frameworks that can transfer bot-relevant knowledge across isolated platforms without sharing raw data.

Federated learning (FL) has emerged as a promising paradigm for training models collaboratively across platforms without exposing local data.
Recent efforts~\cite{zhu2021data,rasouli2020fedgan,zhang2022fine,zhang2022feddtg} have enhanced FL through data-free approaches such as generative adversarial networks and knowledge distillation to further mitigate privacy risks.
It is worth stating plainly why federation rather than a simpler alternative is the right response here. User-level social data is governed by platform terms of service and by regulations such as GDPR, so pooling it across platforms is not merely inconvenient but unlawful; centralized modelling is therefore treated throughout as an upper bound rather than as a competitor. Transfer learning and domain adaptation likewise assume that some party may hold the source data, and are one-directional. Federation is the only paradigm in which every platform both contributes and benefits while its data never leaves its own infrastructure.

However, existing federated methods face several critical limitations:

\begin{itemize}  [leftmargin=*]
\item \textbf{Rigid, Client-Agnostic Message Passing.}
 Federated graph learning methods personalize which parameters a client keeps or how its updates are aggregated~\cite{xie2021gcfl,baek2023fedpub,fedssp2024,adpfedgnn2025}, but the propagation rule itself stays fixed: every client, and every node within a client, aggregates from its neighbors in the same way. This is a poor fit for cross-platform bot detection, where platforms differ in graph density and in how much of a user's signal sits in structure rather than in profile attributes. There remains a need for frameworks whose clients can specialize \emph{how} they propagate information while still learning collaboratively.

    \item \textbf{Inconsistent Feature Spaces.}
     Existing federated knowledge distillation techniques, while increasingly applied to diverse modalities~\cite{mhpflid2023,roy2025dual3d}, often assume relatively consistent  feature spaces across clients.   In the context of social bot detection across heterogeneous platforms, discrepancies between global and local data distributions  —arising from platform-specific user populations, label distributions, and relational structure—can lead to model drift and misaligned feature representations, ultimately degrading overall performance. Feature alignment among clients is therefore essential.

    \item \textbf{Suboptimal  Parameter Aggregation.} In heterogeneous  settings, the   contribution  of each client  model to the global update   should dynamically adapt based on the relevance and quality of its data.  While research on client aggregation has surpassed simple heuristic methods, many existing approaches still struggle to effectively assign fine-grained aggregation weights at the neuron level, or lack adaptive mechanisms for personalized updates tailored to shifting local data distributions.
 \end{itemize}

 To address these challenges, we propose \FedRio, a Personalized \underline{Fed}erated Social Bot Detection with Cooperative \underline{R}e\underline{i}nf\underline{o}rced Contrastive Adversarial Distillation framework.
 \FedRio  integrates an adaptive graph neural network backbone,
  federated adversarial knowledge distillation, contrastive representation alignment,
  server-side adaptive aggregation, and client-side reinforcement learning to improve
  collaborative bot detection under heterogeneous federated settings.

At the model level, \FedRio  introduces a graph neural network
  backbone with an adaptive message-passing mechanism, together with a  federated
  knowledge distillation architecture based on generative adversarial networks. Within
   this framework, a global generator captures   transferable distributional knowledge
  and conveys it  to each client   without requiring raw data sharing. To mitigate
  feature-space inconsistency and model drift, each client performs  multi-stage
    adversarial learning with two classifiers, including one globally shared classifier
  and one locally customized classifier, while  contrastive learning is   used to align
  optimization trajectories  between local and global models.

   At the optimization level, \FedRio addresses heterogeneous data
  distributions and client contribution variability through adaptive model integration
  on both the server and client sides. On the server, an adaptive parameter
  aggregation mechanism  assigns neuron-level weights   to refine  global model updates.
    On the client, reinforcement learning determines, in each round, the extent to which
  downloaded global parameters are integrated into the local model.

The novelty of \FedRio lies in where personalization acts. Existing work personalizes at one end of the communication step and at the granularity of a whole client; \FedRio personalizes at both ends and below that granularity, since the server admits each client's shared-classifier update coordinate by coordinate while the client decides per round how much of the returned global model to absorb. Because the shared backbone and classifier are combined in parameter space, they must have matching shapes across clients; the heterogeneity \FedRio supports is therefore functional rather than architectural: each node selects its own message-passing mode, and each client keeps a personalized head that is never aggregated. 

We evaluate \FedRio on TwiBot-20~\cite{feng2021twibot} and MGTAB~\cite{shi2023mgtab}, using the relations each dataset provides: the follow graph for TwiBot-20 and the seven annotated interaction relations for MGTAB. Each dataset is divided among ten clients by a Dirichlet distribution whose concentration parameter $\alpha$ controls how unevenly the labels are spread, from clients that each see a near-uniform label mix ($\alpha=1$) to clients that hold almost a single class ($\alpha=0.05$). All experiments are repeated over twenty random seeds to average out the effect of random initialization and partitioning, and every reported checkpoint is selected on validation accuracy.
Experimental results demonstrate that \FedRio outperforms the strongest federated baseline in six of the eight dataset partition settings, by $1.4$ to $7.4$ macro-F1 points (one-sided paired $t$-test, $p<0.05$), with the margin growing monotonically on both benchmarks as the labels are spread more unevenly across clients; the exceptions are the two most balanced TwiBot-20 settings, where it is level with the
strongest baselines. Our ablations show that the two proposed modules account for this advantage only in part: removing either module changes nothing a paired test can detect when labels are spread evenly, both contribute at intermediate levels of imbalance, and on the two most uneven MGTAB partitions the cooperative encoder is what carries the gain.

The main contributions of this work are summarized as follows:
\begin{itemize}  [leftmargin=*]
\item An adaptive message-passing backbone in which every node chooses how it propagates, so that clients whose graphs differ in density and homophily can specialize how much structure they use instead of sharing a fixed scheme.
    \item A multi-stage adversarial learning and federated knowledge distillation framework is proposed to better transfer global knowledge in heterogeneous federated settings with skewed  data distributions.
    \item A client-side controller trained by reinforcement learning is introduced to set, for each client and communication round, how much of the downloaded global model is absorbed, instead of using a fixed mixing rate.
    \item An adaptive server-side parameter aggregation mechanism is developed to enable fine-grained, neuron-level model integration.
\item Extensive experiments on two benchmark bot detection datasets, repeated over twenty random seeds with validation-selected checkpoints, show that \FedRio matches or outperforms representative federated learning approaches, with the largest gains and the smallest degradation as the label distribution across clients becomes more uneven.
\end{itemize}

The remainder of this paper is organized as follows. Section~\ref{fedrio:literatures} reviews graph-based bot detection and federated knowledge distillation. Section~\ref{fedrio:formulation} formalizes the federated detection problem and fixes notation. Section~\ref{fedrio:method} presents \FedRio, its adaptive message-passing backbone, its distillation and alignment objectives, and its two personalization mechanisms. Section~\ref{fedrio:experiments} reports the experiments, organized around the seven questions listed at its head. Section~\ref{fedrio:conclusion} concludes.

%% file: 2-literatures.tex
\section{Related Works}
 \label{fedrio:literatures}
 In this section, we introduce three aspects of research, social bot detection based on graph neural networks, federated knowledge distillation, and personalized federated learning.

\vspace*{-3pt}
\subsection{Social Bot Detection with Graph Neural Networks}
Social networks carry rich contextual signals such as social familiarity~\cite{dey2018assessing,dey2019assessing}, affiliation similarity~\cite{peng2018anomalous}, and user interactions~\cite{viswanath2009evolution}, which makes their graph structure well suited to countering bots' mimicry and adaptive evolution with graph neural networks (GNNs). Early detectors combined GNNs with multilayer perceptrons and belief propagation~\cite{ali2019detect}; later work built heterogeneous graphs over features extracted by pre-trained language models and aggregated them with relational GCNs~\cite{feng2021botrgcn} or relation-aware graph transformers~\cite{feng2022heterogeneity}, or fused graph and textual signals~\cite{guo2021social,lei2022bic}. More recent detectors strengthen semantics, self-supervision and robustness: ETS-MM~\cite{etsmm2024} synergizes language models with GNNs, LMBot~\cite{cai2024lmbot} distils graph knowledge into a language model for graph-less deployment, BotTrans~\cite{bottrans2023} transfers across networks by domain adaptation, CACL~\cite{cacl2024} and SATAR~\cite{feng2021satar} use contrastive and self-supervised tuning to track evolving bot communities, SEBOT~\cite{yang2024sebot} adds structural-entropy-guided multi-view contrast, dynamic graph transformers~\cite{he2024dynamicity} capture temporal drift, and RoBCtrl~\cite{yang2025robctrl} shows that reinforced manipulation of bot interactions can evade such detectors.

These methods are centralized, and the assumptions behind their gains do not survive federation. The gains of ETS-MM and LMBot come from optimizing a language model over raw posts---ETS-MM jointly with the GNN, LMBot as the student that replaces the graph at deployment. In our setting text is available only as a per-account embedding computed once and then frozen, so neither mechanism has anything to act on. BotTrans needs a labeled source network and the target network to be visible at the same time, which is precisely what federation forbids. CACL and BotMoE~\cite{liu2023botmoe} operate on a platform-wide social graph: our benchmarks do carry the platforms' own relations, but each client observes only the subgraph induced on the accounts it holds, and restoring the cross-client edges would disclose exactly what the protocol protects. Porting any of them would therefore mean replacing the component that carries most of their published benefit. \FedRio instead integrates an adaptive message-passing mechanism with a federated protocol, collaboratively distilling global knowledge without raw data sharing and thereby improving resilience to evolving bot strategies across heterogeneous platforms.

\vspace*{-3pt}
\subsection{Federated Knowledge Distillation}
Knowledge distillation transfers a teacher's output distribution to a student~\cite{bucilu2006model,hinton2015distilling}, which suits federated learning (FL) because knowledge can move between clients without their data ever moving. Federated variants differ mainly in how the teacher is formed: from client logits refined by forward passes~\cite{seo2020federated}, from an ensemble of local models~\cite{lin2020ensemble}, or from a global generator trained on averaged~\cite{zhu2021data} or per-client~\cite{zhang2022fine} logits. All of them distil through a single global teacher and do not adapt what each client absorbs, so a client whose label distribution diverges sharply from the global mixture gains little from the transfer. FedACK~\cite{yang2023fedack} adds GAN-based bidirectional distillation and was reported to reach state-of-the-art results on heterogeneous data, while MH-pFLID~\cite{mhpflid2023} and DUAL3D-Fed~\cite{roy2025dual3d} carry federated distillation into medical and 3D continual settings. We reimplement FedACK from its published description and run it under our own protocol rather than quoting numbers obtained under a different one. Federated KD nevertheless remains largely unexplored for social bot detection, which is still addressed with isolated per-platform models~\cite{cresci2020decade} even though bots operate across several platforms at once.

\vspace*{-3pt}
\subsection{Personalized Federated Learning}
Personalized FL trains models that generalize to each client's own test distribution, either by personalizing parameters under a shared architecture or by personalizing the architecture itself. At the parameter level, meta-learning~\cite{fallah2020personalized} and Moreau-envelope regularization~\cite{t2020personalized} produce a global model that adapts quickly to local data, local fine-tuning specializes it after aggregation~\cite{wang2019federated,arivazhagan2019federated,yu2020salvaging}, and hybrid schemes interpolate between global and local models or search for their best weighted combination~\cite{mansour2020three,deng2020adaptive,zhang2020personalized}; all of them remain anchored to the global model, which limits local customization. At the architecture level, clients are clustered and served one model per group~\cite{huang2021personalized,ghosh2020efficient}, of which FedAMP~\cite{huang2021personalized} is the limiting case of one personalized model per client, or the unified global model is abandoned altogether, as in the multi-task formulation of MOCHA~\cite{smith2017federated} and the hypernetwork of FedHN~\cite{zhang2021parameterized}. On graph data, FedSSP~\cite{fedssp2024} shares generic spectral knowledge while keeping personalized preference modules and ADPFedGNN~\cite{adpfedgnn2025} decouples global from local parameters by minimizing mutual information, whereas GCFL+~\cite{xie2021gcfl} groups clients by the dynamic time warping distance between their gradient-norm sequences and FED-PUB~\cite{baek2023fedpub} compares functional embeddings obtained by probing each client model with a shared random graph. The last two personalize the aggregation itself rather than only the magnitude of each client's update, so we adopt them as baselines; GCFL+ was developed for graph classification and we apply its clustering mechanism to the node-classification setting studied here. Personalization remains difficult for architectures as complex as GNNs, and \FedRio addresses it by adapting local parameter updates and aggregating global parameters through learned mechanisms tailored to cross-platform bot detection.

%% file: 3-preliminaries.tex
\section{Problem Formulation and Notation}
\label{fedrio:formulation}
In the federated learning setting for social bot detection, the system consists of a central server and \textit{K} clients, each holding a private dataset $\{\mathcal{D}_1, \dots, \mathcal{D}_K\}$. These datasets contain a mixture of benign user accounts and social bots from various generations, and they differ across clients both in label composition and in the structure of the local graph, which is the heterogeneity the method has to absorb.
To address this challenge, the proposed method employs reinforcement learning to adaptively adjust aggregation parameters during the local update process at each client. Specifically, each client $k$ maintains a local model composed of a GNN backbone and a fully connected layer. The GNN dynamically adjusts the message propagation for each node to enable personalized node representations. The overall objective is to minimize the total error across all clients.
The server does not collect raw data from clients but aggregates model parameters to tackle challenges arising from non-identically distributed data. During global aggregation, the server learns a unique aggregation parameter for each client to perform weighted averaging. The overall optimization objective of the proposed method is to minimize the following global loss function:
\begin{equation}
    \arg\min \limits_{w} \mathcal{L}(w) = \frac{1}{K} \sum\limits_{k=1}^K \frac{1}{N_k} \sum\limits_{i=1}^{N_k} \mathcal{L}(x^k_i, y^k_i; w),
    \label{lossCE}
\end{equation}
where $\mathcal{L}$ denotes the loss function used to evaluate the predictive model $w$ on client $k$'s data sample $(x^k_i, y^k_i)$, and $\mathcal{D}_k = \{(x^k_i, y^k_i)\}_{i=1}^{N_k}$.

%% file: 4-methodology.tex
 \begin{figure*}[t]
\centering
\includegraphics[width=0.85\textwidth]{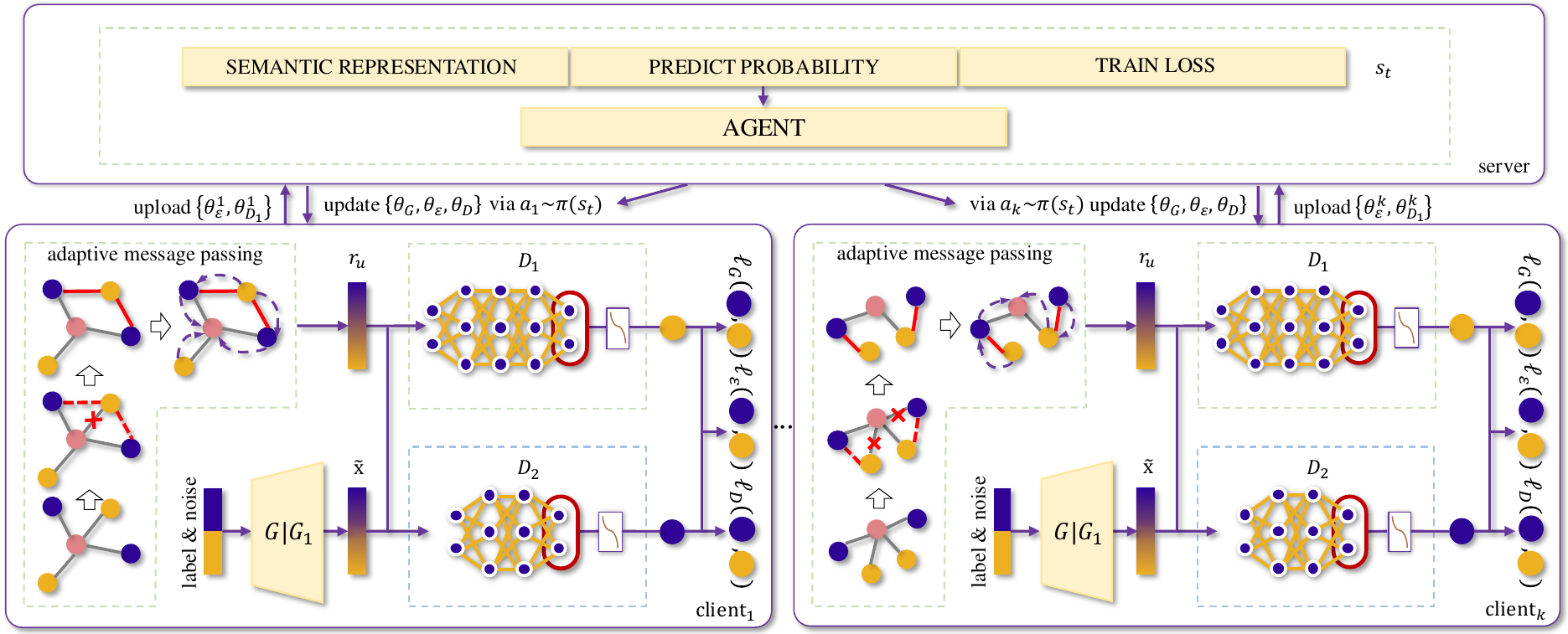}
\vspace{-0.8em}
\caption{The proposed \FedRio framework.}
\label{fig:framework}
\end{figure*}
 \section{Method}
\label{fedrio:method}
\vspace*{-3pt}
\subsection{Overview}
As illustrated in Figure~\ref{fig:framework}, \FedRio comprises four key components: an adaptive message passing module as the backbone model, a federated adversarial contrastive knowledge distillation mechanism, a client-driven reinforcement learning-based parameter update module, and a server-side adaptive parameter aggregation module.
The \textbf{adaptive message passing module} is designed to dynamically adjust the message propagation and aggregation process within the backbone model, enabling platform-specific personalization to address distributional differences across social media platforms. 
The \textbf{federated adversarial contrastive knowledge distillation module} facilitates the sharing of data distribution knowledge across platforms without requiring data exchange. It employs a multi-stage adversarial training mechanism to align feature spaces across clients, promoting more effective knowledge sharing.
The third component is the \textbf{client-side reinforcement learning module}, which determines an appropriate parameter update rate during local model optimization.
The fourth component is the \textbf{server-side adaptive parameter aggregation module}, which learns individualized aggregation weights for each client to optimize the global model updates.

 \subsubsection{Adaptive Message Passing Module}
The adaptive message passing module is deployed within the backbone GNN model $\varepsilon$ of each client. It comprises a pair of cooperative networks: the action network and the environment network, implemented using GraphSAGE and GIN, respectively. The action network aggregates node features and outputs a probability distribution over actions (i.e., message passing modes) for each node. These actions guide how the environment network, centered on a target node, propagates and filters messages from its neighbors during representation learning. This design enables node-level personalized representations.

\vspace*{-2pt}
\subsubsection{Federated Adversarial Contrastive Knowledge Distillation Mechanism}
The proposed framework follows a standard server–client paradigm, where each client maintains a global generator $G$ for knowledge distillation and a local generator $G_k$ for data augmentation. Two classifiers, $D_1$ and $D_2$, are used: classifier $D_1$ has a consistent architecture and parameter initialization across clients but is trained independently on each local dataset; classifier $D_2$ is client-specific and may differ in structure to accommodate model customization needs.

The local dataset $\mathcal{D}_k$ is first encoded by the adaptive message passing backbone model $\varepsilon$ to obtain user representations $r_u = \varepsilon(x_u)$. A multi-stage adversarial training mechanism is then applied, in which the global generator $G$ serves as a medium for transferring data distribution knowledge across clients without sharing raw data. Simultaneously, both classifiers within each client are jointly optimized, and contrastive learning is employed to align heterogeneous feature spaces across clients.

\vspace*{-2pt}
\subsubsection{Client-Guided Reinforcement Learning-Based Parameter Update Module}
\label{sec:privacy}
The adaptive parameter update process in each client's backbone model is formulated as a reinforcement learning (RL)-based policy optimization problem. At the beginning of each communication round, an RL agent determines the update momentum for each client. This federated optimization process is modeled as a Markov Decision Process (MDP), where the state \(s\) is constructed by concatenating the representation vectors of aggregated client nodes, the predicted label distributions, and the client-specific loss values.

Given a state, the RL agent selects an action \(a\) to determine the weighted contribution of the global model parameters when updating the local model. Afterward, the agent receives a reward signal \(r\) based on the accuracy of the aggregated global model on a held-out validation set, enabling it to improve its policy. No raw records leave a client: what is uploaded each round is the model parameters together with the state the controller conditions on, namely a pooled representation, the predicted class probabilities, and a scalar loss. These summaries are nevertheless derived from client data, and parameter and representation exchange is known to be vulnerable to inversion and membership inference attacks, so \FedRio provides no formal privacy guarantee in its default configuration. For deployments that require one, the controller state can be clipped and perturbed with Gaussian noise before upload; Section~\ref{exp:sensitivity} reports how detection quality varies with the noise level.

\vspace*{-2pt}
\subsubsection{Server-Side Adaptive Parameter Aggregation Module}
Under non-IID data distributions, each client's data represents only a small fraction of the overall global distribution, and thus contributes differently to the final global model. To account for this, the server learns a trainable parameter mask for each client. At each aggregation step, the shared classifier parameters uploaded by a client are multiplied element-wise by that client's mask before being averaged into the global model, while the backbone is combined by ordinary sample-weighted averaging. Fusion is thus adaptive at the level of individual neurons.

\vspace*{-3pt}
\subsection{Adaptive Message Passing Module}

The backbone model in this work operates on graph-structured data, where edges between nodes represent the follow or interaction relations that the platform itself records among user accounts. Each node is associated with a set of initial features combining numerical and categorical account metadata with a fixed sentence embedding of the account's textual content, computed once by a pre-trained language model and never updated during federated training.

Federation changes what this graph looks like from inside a client. Each client holds a disjoint set of accounts and can see only those relations whose two endpoints both fall inside that set; a relation between an account it holds and an account held by another client is invisible to both and enters neither computation graph. The density and label composition of the fragment a client sees are determined by how accounts were partitioned, and the adaptive message passing module described below acts downstream of that: it decides which of a client's observable relations carry messages at each layer, not which relations exist. Partitioning therefore induces two kinds of heterogeneity at once: the label skew that federated learning usually studies, and a loss of connectivity that has no counterpart outside graph-structured data.

Due to the significant differences in graph data distributions across platforms, the performance of a model trained on one platform often fails to generalize to others. This is because different spatial-based GNN architectures essentially differ in how they control message passing and aggregation. In a traditional GNN, all nodes share the same message propagation and aggregation scheme, which limits its flexibility in handling heterogeneous data. Applying a uniform message passing mechanism on one platform may result in suboptimal detection performance on another.

To address this challenge, this work adopts a collaborative adaptive message passing module as the backbone model on each client. The key motivation lies in its ability to dynamically adjust the message propagation mode for each node and adaptively aggregate information to generate personalized node representations. Consequently, the learned strategies for controlling node-level message propagation and bot detection patterns can be effectively shared across platforms. This flexibility in adjusting message-passing behaviors enhances the model's ability to adapt to diverse platform-specific data characteristics, resulting in more accurate and efficient social bot detection, thereby improving the overall system effectiveness.

The adaptive message passing mechanism proceeds as follows. Given input node features $x$ (omitting node indices) composed into a matrix $\mathbf{X}$, the first step is to apply layer normalization to stabilize and accelerate training:
\begin{equation}
h = \lambda \cdot \frac{x - E(x)}{\sqrt{Var(x)} + \delta} + \beta,
\end{equation}
where $h \in \mathbf{H}$ denotes the normalized node feature matrix and $\delta$ is a small constant added for numerical stability. The normalized features $\mathbf{H}$ and adjacency matrix $A$ are then fed into two separate but structurally identical GraphSAGE networks: one controls message reception ($U_{in}$), and the other controls message emission ($U_{out}$). These GNNs output a probability distribution over actions for each node $v$:
\begin{equation}
p^{v,l}_{in} = U_{in}^{l}(h_{v}^{l},  M_{in}^{l}\{h_{u}^{l} | u \in N(v)\}),
\end{equation}
\begin{equation}
p^{v,l}_{out} = U_{out}^{l}(h_{v}^{l}, M_{out}^{l}\{h_{u}^{l} | u \in N(v)\}),
\end{equation}
where  $M_{in}^{l}$ and $M_{out}^{l}$ denote the neighborhood aggregation functions (e.g., MEAN, MAX, or SUM) utilized at layer $l$ to collect features from node $v$'s immediate neighbors $N(v)$. Each network outputs a distribution over the action space $\Omega = \{\textsc{keep}, \textsc{drop}\}$, so $p^{v,l}_{in}$ and $p^{v,l}_{out}$ are two-dimensional: at every layer, node $v$ decides independently whether to accept incoming messages and whether to emit outgoing ones. The two decisions combine into the four message-passing modes available to a node at that layer, since it may participate fully, only listen, only broadcast, or isolate itself. Where the layer is clear from context we omit the superscript $l$.

To turn these distributions into discrete edge decisions that can still be trained end to end, we use the straight-through Gumbel-Softmax estimator. From i.i.d. Gumbel noise,
\begin{equation}
g_i^v = -\log(-\log(\epsilon_i^v)),
\label{fedrio:eq:g_noise}
\end{equation}
a relaxed sample is drawn,
\begin{equation}
\tilde{y}_i^v = \frac{\exp((\log(p_i^v) + g_i^v)/\tau)}{\sum_{j=1}^{|\Omega|} \exp((\log(p_j^v) + g_j^v)/\tau)},
\end{equation}
and is then hardened into a one-hot decision in the forward pass,
\begin{equation}
    GS(p^v)_i = \mathbf{1}\left[\, i = \arg\max\nolimits_{j}\, \tilde{y}_j^v \,\right],
\label{fedrio:eq:gs_hard}
\end{equation}
where $\epsilon_i^v \sim \textit{Uniform}(0, 1)$ is a uniformly distributed random variable, $g_i^v$ is an i.i.d. Gumbel noise term, $p_i^v$ is the probability of action $a_i \in \Omega$ for node $v$, and $\tau$ is the softmax temperature. In the backward pass the gradient of the relaxed sample $\tilde{y}^v$ is used in place of that of the hardened one, so the discrete choice remains trainable. Since $GS(\cdot)$ returns a one-hot vector rather than a probability vector, the edge decisions defined below are genuinely discrete rather than soft weights.

Writing $GS(\cdot)_{\textsc{keep}}$ for the coordinate of the sampled action that corresponds to \textsc{keep}, the gate on the edge from $u$ to $v$ is
\begin{equation}
w_{uv} = GS(p^u_{out})_{\textsc{keep}} \cdot GS(p^v_{in})_{\textsc{keep}} \in \{0, 1\},
\end{equation}
so an edge carries a message at this layer only when the source node $u$ chooses to emit and the target node $v$ chooses to receive; in every other case the gate is exactly zero.

Because the gate takes values in $\{0,1\}$, multiplying the message on an edge by it is equivalent to deleting that edge from the layer's computation graph:
\begin{equation}
A'_{uv} = A_{uv} \cdot w_{uv}.
\end{equation}
The fraction of edges that survive a layer is then the mean of $w_{uv}$ over the edges of $A$, a measured outcome of training rather than a threshold we set; we report it with the module validation results in Section~\ref{sec:module-validation}.

The updated adjacency matrix $A'$ and the node representations are then fed into the environment network $U_{env}$ (implemented using GIN) to generate updated node embeddings:
\begin{equation}
h^{l+1}_{v} = h^{l}_{v} + U_{env}^{l}\left(h_{v}^{l},\ M_{env}^{l}\{h_{u}^{l} \mid u \in N'(v)\}\right),
\end{equation}
where  $M_{env}^{l}$ denotes the neighborhood aggregation function in the environment network, $N'(v)$ denotes the neighbors of $v$ under the updated adjacency matrix $A'$, and $h^{l+1}_{v}$ is the representation of node $v$ after layer $l$. The layer output is added to the layer input rather than replacing it, and layer normalization is applied to the input of every layer. Both are standard stabilization choices, and both are used identically in the matched-capacity backbone against which this module is compared, so that the comparison isolates the effect of the gates.

Finally, the last-layer embedding $h^{L}_{v}$ is passed through a fully connected layer to produce the final output representation $r_v$ from the backbone model, which is used in subsequent training stages.

\vspace*{-3pt}
\subsection{Client-Driven Federated Adversarial Contrastive Knowledge Distillation}

\vspace*{-2pt}
\subsubsection{Local Adversarial Contrastive Knowledge Distillation}
Each client adopts a multi-stage adversarial contrastive knowledge distillation strategy to address the challenge of non-IID data distributions across clients.

\mypara{Stage 1: Training classifiers $D_1$ and $D_2$.}
This stage aims to encourage the model to learn distinct decision boundaries for the same class and to compress the feature space generated by the backbone model. Given the \textit{non-IID} nature and \textit{data scarcity} on clients, the global generator $G$ acts as a teacher network, providing knowledge of the global data distribution. For a sample $(x^k_i, y^k_i)$, $G$ synthesizes a pseudo-sample $\tilde{x} = G(z, y^k_i; \theta_G)$, where $z \sim \mathcal{N}(0,1)$ is a standard Gaussian noise and $y^k_i$ is the label. The classifier $D_1$ then processes both $(x^k_i, \tilde{x})$ to obtain the predicted distributions $(p, \tilde{p})$. The distillation loss minimizes the divergence between the probability distribution $p$ of real data $x^k_i$ and that of the synthetic data $\tilde{p}$, as defined in Eq.~(\ref{fedrio:eq:distill}):
\begin{equation}
\mathcal{L}^k_{dis} = \frac{1}{N_k}\sum\limits^{N_k}_{i=1}
    D_{KL}(\sigma(D_1(r^k_i))\|\sigma(D_1(\tilde{x}))), \label{fedrio:eq:distill}
\end{equation}
where $\sigma$ is the softmax function, $D_{KL}$ denotes the Kullback-Leibler divergence, and $r^k_i = \varepsilon(x^k_i)$ is the feature representation produced by the backbone model. A similar loss ${\mathcal{L}^k_{dis}}'$ is computed for classifier $D_2$ using the same formula.

To further refine the feature space, we introduce an \textit{adversarial loss} that measures the divergence between the output distributions of $D_1$ and $D_2$ for the same input:
\begin{equation}
\mathcal{L}^k_{adv} = \frac{1}{N_k}\sum\limits^{N_k}_{i=1} D_{KL}(\sigma(D_1(r^k_i))\|\sigma(D_2(r^k_i))). \label{fedrio:eq:lossadv}
\end{equation}
Maximizing Eq.~(\ref{fedrio:eq:lossadv}) encourages $D_1$ and $D_2$ to learn diverse decision boundaries, thus promoting a more granular partitioning of the feature space. Intuitively, if the feature vectors produced by $\varepsilon$ can be correctly classified by both classifiers despite differing boundaries, then the vectors must lie in their overlapping classification region, leading to finer feature learning. Similarly, we apply Eq.~(\ref{fedrio:eq:lossadv}) to the pseudo-data generated by the local generator $G_k$, computing an adversarial loss $\mathcal{L}^k_{advg}$, which is minimized.
In addition, $D_1$ and $D_2$ are required to correctly classify pseudo-data randomly generated by $G$, helping to address data imbalance and scarcity. In summary, the total loss for training both classifiers is defined as:
\begin{equation}
\mathcal{L}^k_D = \mathcal{L}^k_{cls} + \alpha_{dis}(\mathcal{L}^k_{dis}+{\mathcal{L}^k_{dis}}') + \gamma_{adv}(\mathcal{L}^k_{advg} -\mathcal{L}^k_{adv}), \label{fedrio:eq:lossD}
\end{equation}
where $\alpha_{dis}$ and $\gamma_{adv}$ weight the distillation and adversarial terms.

\mypara{Stage 2: Training the backbone model $\varepsilon$.}
This stage begins by minimizing the adversarial loss in Eq.~(\ref{fedrio:eq:lossadv}) to reduce the divergence between the output distributions of the two classifiers for the same feature vector. A lower divergence implies that the data point lies on the same side of both decision boundaries, thereby encouraging the backbone model to produce more precise and compact feature representations.

Contrastive learning is employed to guide the parameter optimization of the backbone model, mitigating \textit{model drift} between local and global extractors. At the $t$-th communication round, the objective is to ensure that the representation $r = \varepsilon^k_{t}(x^k_i)$ produced by the client-side backbone $\varepsilon^k_{t}$ aligns closely with the global representation $r_{glo} = \varepsilon_{t}(x^k_i)$, while remaining dissimilar from the previous round’s representation $r_{pre} = \varepsilon_{t-1}(x^k_i)$. The contrastive loss is defined as:
\begin{equation}
    \mathcal{L}^{k,i}_{con} = -\log \frac{\exp({\rm sim}(r, r_{glo})/\tau_{con})}{\exp({\rm sim}(r, r_{glo})/\tau_{con}) + \exp({\rm sim}(r, r_{pre})/\tau_{con})}, \label{fedrio:eq:losscon}
\end{equation}
where ${\rm sim}(\cdot, \cdot)$ denotes a similarity function, and $\tau_{con}$ is a temperature scaling factor. This contrastive approach not only alleviates local model drift but also leverages the global model as an adversarial mediator to coordinate the feature spaces across different clients.
Accordingly, the total loss function for training the feature extractor is given by:
\begin{equation}
    \mathcal{L}^k_{\varepsilon} = \mathcal{L}^k_{cls} + \gamma_{adv} \mathcal{L}^k_{adv} + \mu_{con} \frac{1}{N_k} \sum\limits^{N_k}_{i=1} \mathcal{L}^{k,i}_{con}, 
\label{fedrio:eq:losse}
\end{equation}
where $\mu_{con}$ weights the contrastive term.

\mypara{Stage 3: Training $G_k$.}
To ensure that $G_k$ generates pseudo samples lying near the decision boundaries of both classifiers, the adversarial loss $\mathcal{L}^k_{advg}$ is maximized. This encourages the boundaries to converge toward overlapping regions and, consequently, compresses the consistent feature space learned by $\varepsilon$.

Since $G_k$ generates samples conditioned only on class labels, a diversity loss $\mathcal{L}_{var}$ is introduced to enhance the variability of generated samples and prevent mode collapse:
\begin{equation}
\begin{aligned}
    \mathcal{L}_{var} = \exp \left( \frac{1}{N^2} \sum\limits_{i,j \in \{1,\cdots,N\}} 
    \left( -\| \tilde{x}_i - \tilde{x}_j \|_2 \cdot \| z_i - z_j \|_2 \right) \right),
\end{aligned}
\end{equation}
where $\tilde{x}_i = G_k(z_i, \hat{y}_i)$. The overall loss for the generator $G_k$ is defined as:
\begin{equation}
    \mathcal{L}^k_{g} = \mathcal{L}^k_{cls} - \mathcal{L}^k_{advg} + \mathcal{L}_{var}. \label{fedrio:eq:lossgen}
\end{equation}

After local training, each client $k$ transmits the parameters $\{\Theta^k_\varepsilon, \Theta^k_{D_1}\}$ to the server. The server then performs adaptive parameter aggregation, as detailed in Section~\ref{fedrio:method:serveragg}, to update the global parameters of $G$, $D_1$, and $\varepsilon$. Subsequently, each client downloads the updated global models and integrates them into its local environment following the procedure described in Section~\ref{fedrio:method:clientagg}, thereby initiating the next round of client training.

\vspace*{-2pt}
\subsubsection{Server-Side Aggregated Knowledge Extraction}
To enable knowledge sharing of local data distributions across clients, the global generator $G$ is used to extract global knowledge without relying on proxy data for server-side distillation. In this process, the shared classifiers $\{D^k_1\}$ of all clients jointly act as the teacher: the generator is trained so that every client classifier assigns its output to the intended label, weighted by how much of that label the client holds. The loss function for updating $G$ is defined as:
\begin{equation}
\begin{aligned}
    \mathcal{L}_{G} = \frac{1}{K} \sum_{k=1}^{K} \frac{1}{B} \sum_{\tilde{x}} \rho^{k,y}_t \,
    \mathcal{L}_{cls}\left( D^k_1(\tilde{x}),\, y \right) + \lambda_{div}\, \mathcal{L}_{div}, \label{fedrio:eq:lossG}
\end{aligned}
\end{equation}
where $\mathcal{L}_{cls}$ is the cross-entropy loss, $B$ is the number of synthetic samples drawn per step, and $\tilde{x} = G(y, z)$ is a synthetic sample produced by $G$ from noise $z \sim \mathcal{N}(0,\mathbf{I})$ and a label $y$ sampled according to the label distribution $p(y)$; the samples drawn at each step form the experience set $\mathcal{D}_G$. Because the classifiers operate on the output of the backbone, $\tilde{x}$ lives in the representation space rather than the raw node feature space. The coefficient $\rho^{k,y}_t$ denotes the proportion of samples with label $y$ in client $k$ relative to the total number of label-$y$ samples in the global dataset $\mathcal{D}$. This weighting matters under label skew: it draws the teaching signal for a class from the clients that hold most of it, rather than from clients whose view of that class rests on a handful of samples.

The second term guards against mode collapse, which matters here because the aggregation masks are fitted on the generator's own samples (Eq.~\ref{fedrio:eq:lagg}): a generator that emits a single prototype per class would tune them to that prototype rather than to the label distribution. It penalizes outputs that stay close together, weighting each pair by how far apart the noise vectors that produced it are:
\begin{equation}
    \mathcal{L}_{div} = \exp\left( - \frac{1}{B^2}\sum_{i,j} \lVert \tilde{x}_i - \tilde{x}_j \rVert \cdot \lVert z_i - z_j \rVert \right),
    \label{fedrio:eq:lossdiv}
\end{equation}
This is the same criterion already applied to the client generators in Stage 3. We set $\lambda_{div} = 0.1$ in all experiments.

The label distribution $p(y)$ is estimated through the exchange of label statistics among clients during communication rounds. This allows the generator $G$ to acquire knowledge of the global data distribution and enables multi-stage training to propagate such knowledge to each client model effectively.

\vspace*{-3pt}
\subsection{Reinforcement Learning-Based Client Parameter Update}
\label{fedrio:method:clientagg}
\vspace*{-2pt}
\subsubsection{Deep Q-Network-Based Reinforcement Learning}

This module aims to enhance the detection accuracy of social bots on different platforms by learning appropriate parameter update weights for each client using a reinforcement learning (RL) agent. The definitions of state, action, and reward in the RL setting are introduced below.

\mypara{State.}
At each time step, the state \( s_t \) is constructed using the node feature vectors, predicted probability distributions, and corresponding loss values generated by the backbone models of all clients. It is represented as:
\begin{equation}
    s_t = [ \mathbf{r}_t^{1}, \mathbf{p}_t^{1}, \ell_t^{1}, \ldots, \mathbf{r}_t^{K}, \mathbf{p}_t^{K}, \ell_t^{K} ],
\end{equation}
where \( \mathbf{r}_t^{k} \) denotes the feature vector of client \( k \) at communication round \( t \),
\( \mathbf{p}_t^{k} \) denotes the predicted probability distribution of client \( k \) at round \( t \),
and \( \ell_t^{k} \) is the corresponding loss value.

\mypara{Action Selection and Parameter Update.}
At the end of each communication round in federated learning, the RL agent predicts the Q-values for each action under the current state. These Q-values are then converted into a probability distribution using the softmax function, which determines the weights for updating each client's model parameters. The action selection distribution \( \pi(a|s_t) \) is defined as:
\begin{equation}
    \pi(a|s_t) = \frac{\exp(Q(s_t, a))}{\sum_{a'} \exp(Q(s_t, a'))},
    \label{fedrio:eq:pi}
\end{equation}
where \( Q(s_t, a) \) is the Q-value for action \( a \) in state \( s_t \). The optimal action \( a_t \) is selected based on the distribution \( \pi(a|s_t) \).

Each component \( a^k_t \in (0,1) \) of the selected action \( a_t \) is the parameter update weight for client \( k \), and the update rule for that client is:
\begin{equation}
    \Theta^k_{t,0} = (1 - a^k_t) \Theta^k_{t-1} + a^k_t \Theta_{t-1},
    \label{fedrio:eq:updateclient}
\end{equation}
where \( \Theta^k_{t,0} \) denotes the parameters of client \( k \) at the start of round \( t \), before local training begins, \( \Theta^k_{t-1} \) the parameters it held at the end of the previous communication round, \( \Theta_{t-1} \) the global parameters broadcast by the server, and \( \alpha \) the update weight.

\mypara{Reward Mechanism.}
The reward observed at each time step is defined as \( r_t = \xi^{(\omega_t - \omega^{*})} - 1 \), where \( \omega_t \) denotes the accuracy of the aggregated global model on the held-out validation set at round \( t \), \( \omega^{*} \) is the target accuracy, and \( \xi > 1 \) controls how sharply the reward grows with \( \omega_t \). We set \( \xi = 2 \) and \( \omega^{*} = 0.95 \) in all experiments. As long as \( \omega_t \leq \omega^{*} \), the reward lies in \( (\xi^{-1} - 1,\, 0] \) and attains its maximum of \( 0 \) only at the target.

The objective of the reinforcement learning agent is to maximize the expected cumulative discounted reward, defined as:
\begin{equation}
    R = \sum_{t=1}^{T} \gamma^{t-1} r_t = \sum_{t=1}^{T} \gamma^{t-1}(\xi ^{(\omega_t - \omega^{*})} - 1), 
    \label{fedrio:eq:reward}
\end{equation}
where \( \gamma \in (0,1) \) is the discount factor, which controls the agent's preference for immediate versus long-term rewards.

As training progresses, improvements in accuracy tend to slow down. Since the derivative of \( \xi^{(\omega_t - \omega^{*})} \) grows as \( \omega_t \) approaches \( \omega^{*} \), a gain of a given size is worth more late in training than early, which keeps the incentive meaningful in the later rounds. The constant term \( -1 \) makes every round short of the target contribute negatively, so the agent is pushed to reach the target accuracy in fewer communication rounds.

\vspace*{-2pt}
\subsubsection{Reinforcement Learning Optimization}
We employ the Double Deep Q Network (DDQN) to train a neural network that approximates the optimal action-value function \( Q^*(s_t, a) \). The value function provides an estimate of the expected return for each possible action \( a \) in a given state \( s_t \). The original Q-learning algorithm is known to suffer from instability because it indirectly optimizes the reinforcement learning agent by learning an approximator \( Q(s, a; \theta_t) \).
DDQN stabilizes the estimation of the action-value function by introducing a second value function \( Q(s, a; \theta_t') \). To train the reinforcement learning agent, parameter aggregation weights for clients are randomly selected during the first \( t \) communication rounds. The state, action, and reward tuples obtained from the environment are stored.

After several rounds of federated learning communication, the training of the reinforcement learning agent begins. Samples of state-action pairs are drawn from the stored experiences, and the network parameters are optimized by minimizing the following loss function:
\begin{equation}
    \ell_t(\theta_t) = \left( Y_t^{\text{DoubleQ}} - Q(s_t, a; \theta_t) \right)^2,
    \label{fedrio:eq:lossRL}
\end{equation}
where \( Y_t^{\text{DoubleQ}} \) denotes the target value at round \( t \), defined as:
\begin{equation}
    Y_t^{\text{DoubleQ}} = r_t + \gamma Q\left(s_{t+1}, \underset{a}{\arg\max} Q(s_{t+1}, a; \theta_t); \theta_t'\right),
    \label{fedrio:eq:ddqn}
\end{equation}
where the target \( Y_t^{\text{DoubleQ}} \) is updated using two action-value functions, where \( \theta_t \) denotes the online network parameters updated at each step and \( \theta_t' \) represents the frozen target network parameters that provide stability for the action-value estimation. The action-value function \( Q(s_{t+1}, a; \theta_t) \) is updated by minimizing the loss \( \ell_t(\theta_t) \) via gradient descent: 

\begin{equation}
    \theta_{t+1} = \theta_t + \eta \left( Y_t^{\text{DoubleQ}} - Q(s_t, a; \theta_t) \right) \nabla_{\theta_t} Q(s_t, a; \theta_t),
    \label{fedrio:eq:updateqnetwork}
\end{equation}
where \( \eta \) is the learning rate and \( \nabla_{\theta_t} Q(s_t, a; \theta_t) \) denotes the gradient of \( Q \) with respect to \( \theta_t \).

\vspace*{-3pt}
\subsection{Server-Side Adaptive Parameter Aggregation}
\label{fedrio:method:serveragg}

To account for the varying importance of different client models in aggregating global model parameters, this section proposes an adaptive parameter aggregation method. The method performs fine-grained, neuron-level integration of global and local model parameters. At the start of federated training, the server initializes for each client \( k \) a learnable aggregation weight matrix \( W_k \) that has the same shape as the shared classifier's parameters. All entries start at the same value, so that after the normalization of Eq.~\ref{fedrio:eq:masknorm} every client carries weight \( 1/K \) at every coordinate and the first aggregation reduces to plain averaging.

\begin{algorithm}[t]
\small
\caption{\FedRio}\label{fedrio:algorithm1}
\KwIn{
Client datasets $\mathcal{D}_k$, $k=1,\ldots,K$; client generators $G_k$; $\varepsilon$; classifiers $\{D^k_1\}$, $\{D^k_2\}$; global generator $G$
}
\KwOut{Personalized client models $\{D^k_2\}$ and global parameters $\{\Theta_\varepsilon, \Theta_{D_1}, \Theta_G\}$}
Initialize all parameters\;

\For{each communication round $t=1,\ldots,T$}{
    The server broadcasts all model parameters $\{\Theta_G, \Theta_\varepsilon, \Theta_{D_1}\}$ to clients\;\label{fedrio:algl:broadcast}
    \For{each client $k$ \textbf{in parallel}}{
        Clients blend the downloaded global parameters into their local models using Eq.~\ref{fedrio:eq:updateclient}\;
        \For{each training epoch $e=1\sim E$ \tcp*{\textbf{Stage 1}} \label{fedrio:alg:trainD}}{
            Compute classifier loss $\mathcal{L}^k_D$ with Eq.~\ref{fedrio:eq:lossD}\;
            Update classifier parameters: $\{\Theta^k_{D_1}, \Theta^k_{D_2}\} \gets \{\Theta^k_{D_1}, \Theta^k_{D_2}\} - \eta \nabla \mathcal{L}^k_D$\;
        }
        \For{each training epoch $e=1\sim E$ \tcp*{\textbf{Stage 2}} \label{fedrio:trainE}}{
            Compute backbone model loss $\mathcal{L}^k_{\varepsilon}$ via Eq.~\ref{fedrio:eq:losse}\;
            Update backbone parameters: $\Theta^k_\varepsilon \gets \Theta^k_\varepsilon - \eta \nabla \mathcal{L}^k_{\varepsilon}$\;
        }
        \For{each training epoch $e=1\sim E$ \tcp*{\textbf{Stage 3}} \label{fedrio:alg:trainlocalG}}{
            Compute generator loss $\mathcal{L}^k_g$ using Eq.~\ref{fedrio:eq:lossgen}\;
            Update client generator parameters: $\Theta^k_{G_k} \gets \Theta^k_{G_k} - \eta \nabla \mathcal{L}^k_g$\;
        }\label{fedrio:alg:trainGk}
        Client $k$ sends $\{\Theta^k_\varepsilon, \Theta^k_{D_1}\}$ to the server\;
    }
    Aggregate server model parameters using Algorithm~S2 in the appendix\;\label{fedrio:alg:updateserver}
    Update global generator $G$ with Eq.~\ref{fedrio:eq:lossG}\;\label{fedrio:alg:trainG}
    Update reinforcement learning agent via Algorithm~S1 in the appendix\;\label{fedrio:alg:trainRL}
}

\end{algorithm}

At the \( t \)-th communication round of federated learning, each client \( k \) uploads its backbone parameters \( \Theta^k_\varepsilon \) and its shared classifier parameters, which the server aggregates by different rules. The backbone is combined by ordinary sample-weighted averaging, \( \Theta_\varepsilon \gets \sum_k (n_k / \sum_j n_j)\, \Theta^k_\varepsilon \), where \( n_k \) is the number of nodes held by client \( k \). Adaptive aggregation is reserved for the shared classifier. Writing \( \Theta_t^k \) for its parameters at client \( k \) after local training and \( \Theta_{t-1} \) for the global parameters produced at the end of the previous round, the server applies the masked client updates on top of \( \Theta_{t-1} \) to produce the new global parameters \( \Theta_t \):
\begin{equation}
    \Theta_t = \Theta_{t-1} + \sum_k (\Theta_t^k - \Theta_{t-1}) \odot W_k,
    \label{fedrio:eq:aggTheta}
\end{equation}
Every client's update is measured against the same \( \Theta_{t-1} \) rather than against that client's own parameters at the start of the round, which by Eq.~\eqref{fedrio:eq:updateclient} need not coincide with it: a client whose action \( \alpha \) falls below one begins the round from a mixture of the broadcast global model and its own previous one. Sharing the reference across clients is what makes the rule a convex combination of the client models; per-client references would not cancel.

The weight matrices \( W_k \) are obtained from unconstrained mask logits \( m_k \), which are normalized across clients at every parameter coordinate \( i \) by a softmax,
\begin{equation}
    W_k[i] = \frac{\exp(m_k[i])}{\sum_{j=1}^{K} \exp(m_j[i])},
    \label{fedrio:eq:masknorm}
\end{equation}
so that \( W_k[i] > 0 \) and \( \sum_k W_k[i] = 1 \) for every \( i \). Aggregation is therefore a convex combination of the client models: substituting \( \sum_k W_k = 1 \) reduces Eq.~\ref{fedrio:eq:aggTheta} to the weighted average \( \sum_k W_k \odot \Theta_t^k \). We keep the incremental form because it matches the server's computation and makes the per-client update \( \Theta_t^k - \Theta_{t-1} \) explicit.

To optimize the weight matrices, the global generator \( G \) is used to produce synthetic samples. The generator \( G \) takes as input the label vector \( \mathbf{y} \) and Gaussian noise \( \mathbf{z} \) and generates outputs \( \tilde{x} \):
\begin{equation}
    \tilde{x} = G(\mathbf{y}, \mathbf{z}, \Theta_G),
    \label{fedrio:eq:genx}
\end{equation}
where \( \tilde{x} \) denotes the generated synthetic data. At every mask step the server draws a fresh batch of \( B \) such samples, forms the candidate global classifier implied by the current masks, and computes its cross-entropy on that batch:
\begin{equation}
\begin{aligned}
    \mathcal{L}_{agg} = \frac{1}{B} \sum_i &\left[ -y_i \log \hat{p}_i - (1 - y_i) \log(1 - \hat{p}_i) \right], \\ \quad &\hat{p}_i = \sigma\!\left(D_1(\tilde{x}_i;\, \Theta_t)\right).
    \label{fedrio:eq:lagg}
\end{aligned}
\end{equation}

Finally, the mask logits are optimized by backpropagating the loss gradient with learning rate \( \eta \):
\begin{equation}
    m_k = m_k - \eta \nabla_{m_k} \mathcal{L}_{agg},
    \label{fedrio:eq:laggback}
\end{equation}
so that the induced weights stay on the simplex of Eq.~\eqref{fedrio:eq:masknorm} throughout the update.

The gradient in Eq.~\eqref{fedrio:eq:laggback} is computed entirely on synthetic samples, so nothing in it prevents the masks from drifting towards a combination that fits the generator's output better than it fits real data. We therefore do not run the update to its last step unconditionally. After each of the \( S \) mask steps the server forms the candidate global classifier implied by the current masks, evaluates its cross-entropy on a small server-side validation set, and keeps the mask state that attains the lowest such loss; the loop stops early once \( \max(3, \lfloor S/5 \rfloor) \) consecutive steps fail to improve it, and the retained state, not the final one, is used in Eq.~\eqref{fedrio:eq:aggTheta}. The validation set is the same held-out split that supplies the reward in Eq.~\eqref{fedrio:eq:reward} and is never used for gradient computation, so the mechanism selects among candidates produced by synthetic data rather than fitting real data. We state this step explicitly because Eq.~\eqref{fedrio:eq:laggback} on its own is not what the framework runs: an implementation that stops at that equation realizes a different module, one whose masks are accountable only to the generator. The guard should be read as a safeguard on an unsupervised inner loop rather than as the source of the module's contribution, which Section~\ref{sec:module-validation} isolates by removing the module itself.

\vspace*{-3pt}
\subsection{Overall Workflow}
Algorithm~\ref{fedrio:algorithm1} outlines the complete workflow of \FedRio. Initially, all model parameters are initialized, followed by alternating phases of knowledge distillation and reinforcement learning-based selection.
In each communication round, \FedRio first broadcasts the latest $G$, $\varepsilon$, and $D_1$ to all clients (Line~\ref{fedrio:algl:broadcast}). Each client then optimizes its local models $D^k_1$, $D^k_2$, $\varepsilon^k$, and $G_k$ on its own data through three training stages (Lines~\ref{fedrio:alg:trainD}-\ref{fedrio:alg:trainGk}). Upon completing parallel optimization, the server aggregates the clients' parameters from the current round to update the global parameters $\Theta_\varepsilon$ and $\Theta_{D_1}$ (Line~\ref{fedrio:alg:updateserver}) and further optimizes the global generator $G$ (Line~\ref{fedrio:alg:trainG}). Finally, the reinforcement learning agent is optimized (Line~\ref{fedrio:alg:trainRL}).

\vspace*{-3pt}

\vspace*{-3pt}
\subsection{Complexity and Practicality Analysis}
The per-round time and communication costs of each component, together with the corresponding notation, are analysed in the appendix; Section~\ref{exp:efficiency} reports the measured counterparts.

%% file: 5-experiment-setup.tex
\section{Experiments}
\label{fedrio:experiments}
The experiments aim to answer the following questions:
\begin{itemize}[leftmargin=*]
    \item \textbf{Q1}. How does \FedRio perform in classification under different data distribution scenarios?
    \item \textbf{Q2}. What does that performance cost in communication rounds, parameters per round and time per round?
    \item \textbf{Q3}. Do clients converge on a consistent feature space, and is the contrastive objective responsible for it?
    \item \textbf{Q4}. How does \FedRio scale with the number of clients, and how sensitive is the controller to noise added to the state it uploads?
    \item \textbf{Q5}. What does each module contribute, and is a learned controller necessary or do simpler adaptive rules suffice?
    \item \textbf{Q6}. Does the advantage survive a federation whose clients come from \emph{independently collected} datasets rather than from resampling one?
    \item \textbf{Q7}. How does \FedRio compare with centralized social bot detectors, which observe the complete graph and all features?
\end{itemize}

\vspace{-0.7em}
\vspace*{-3pt}
\subsection{Experimental Setup}
\label{exp:setup}

\vspace*{-2pt}
\subsubsection{Software and Hardware}
\FedRio is implemented via Python 3.11.15 with PyTorch 2.4.1, CUDA 12.1 and PyTorch Geometric
2.6.1 and runs on NVIDIA H200 GPUs.

\begin{table}
\centering
\caption{Statistics of the datasets. Edges are the relations the platform records, symmetrized and deduplicated. Homophily is the fraction of edges joining two accounts of the same class.}
\label{tab:dataset_statistics}
\resizebox{0.48\textwidth}{!}{
\begin{tabular}{lrrrrrr}
\toprule
Dataset & Nodes & Edges & Bots & Humans & Feat. & Homoph. \\
\midrule
TwiBot-20 & 11,826 & 46,362 & 6,589 & 5,237 & 794 & 0.588 \\
MGTAB     & 10,199 & 902,856 & 2,748 & 7,451 & 788 & 0.750 \\
\bottomrule
\end{tabular}}
\vspace{-3pt}
\end{table}

\vspace*{-2pt}
\subsubsection{Datasets}
We evaluate on TwiBot-20~\cite{feng2021twibot} and MGTAB~\cite{shi2023mgtab}, the two
public bot benchmarks that ship a social graph alongside account metadata and posted
text. TwiBot-20 provides follow relations; MGTAB provides seven annotated relation
types, all of which we use, symmetrized and deduplicated. Node features concatenate
account properties---$26$ on TwiBot-20 and $20$ on MGTAB---with a $768$-dimensional
sentence embedding of the user's recent posts, computed by English BERT and by the
multilingual LaBSE model MGTAB distributes, whose corpus spans $54$ languages. Both use
a stratified $70/10/20$ split into training, validation and test nodes. The pair is
informative rather than redundant: TwiBot-20 is majority-bot, sparse and only mildly
homophilous, while MGTAB is majority-human, two orders of magnitude denser and
considerably more homophilous (Table~\ref{tab:dataset_statistics}).

\vspace*{-2pt}
\subsubsection{Graph construction}
\label{setup:graph}
The edges are the relations each dataset records, so the adaptive message passing module
selects among observed social ties. Federation then removes part of every graph, in a way a centralized study never
encounters: a client can only use the edges whose two endpoints it holds, and every edge
crossing a client boundary is invisible to all participants. Under our partitions this removes most
of the graph: averaged over the same $20$ seeds, the edges surviving inside clients are
$6.6\%$, $8.9\%$, $14.2\%$ and $21.9\%$ of the TwiBot-20 graph at $\alpha=1$, $0.5$,
$0.1$ and $0.05$, and $7.7\%$, $9.9\%$, $14.5\%$ and $25.7\%$ of the MGTAB
graph---sharper skew retaining more, since a partition concentrating one class on one
client concentrates that class's internal ties there too. A federated detector therefore never
sees the graph on which centralized detectors are evaluated, only fragments of it, which
is why the comparison in Section~\ref{sec:discussion} is contextual rather than
head-to-head.

\vspace*{-2pt}
\subsubsection{Data heterogeneity}
\begin{figure*}[t]
\centering
\subfigure[TwiBot-20 ($\alpha=1.0$).]{
\includegraphics[width=0.32\linewidth]{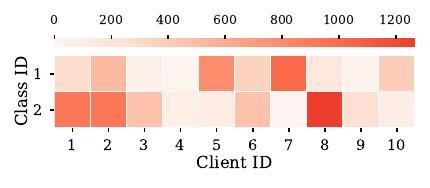}}
\subfigure[TwiBot-20 ($\alpha=0.5$).]{
\includegraphics[width=0.32\linewidth]{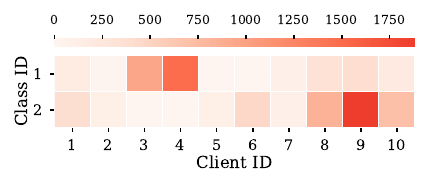}}
\subfigure[TwiBot-20 ($\alpha=0.1$).]{
\includegraphics[width=0.32\linewidth]{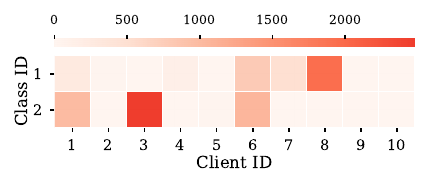}}
\subfigure[MGTAB ($\alpha=1.0$).]{
\includegraphics[width=0.32\linewidth]{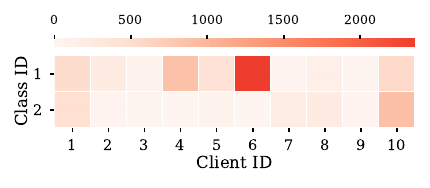}}
\subfigure[MGTAB ($\alpha=0.5$).]{
\includegraphics[width=0.32\linewidth]{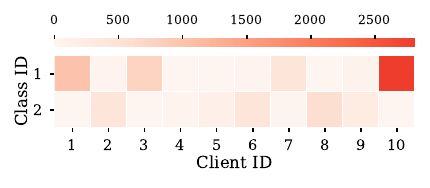}}
\subfigure[MGTAB ($\alpha=0.1$).]{
\includegraphics[width=0.32\linewidth]{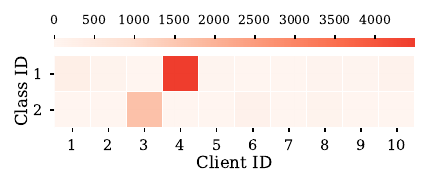}}
\caption{Visualization of data heterogeneity at seed 1: training samples of each class (1 human, 2 bot) held by each client, for three of the four partitions we report. The darker the color, the more samples with that label are available to the client.}
\label{fig:heatmap-datapartition1}
\end{figure*}

Clients receive a Dirichlet $\text{Dir}(\alpha)$ partition of the labeled
accounts~\cite{li2021federated}, with smaller $\alpha$ concentrating each class on
fewer clients, as illustrated in Figure~\ref{fig:heatmap-datapartition1}. We report
$\alpha \in \{1, 0.5, 0.1, 0.05\}$, spanning a near-uniform split and a regime in which
a substantial fraction of clients hold almost a single class. Each client induces its
subgraph on the accounts it receives, as described above.

\vspace*{-2pt}
\subsubsection{Baselines}
Federated distillation is the natural comparison for a framework that shares knowledge
rather than data. We compare against \textbf{FedAvg}~\cite{mcmahan2017communication} and
\textbf{FedProx}~\cite{li2020federated}, which constrain local training under
heterogeneity; \textbf{FedEnsemble} and \textbf{FedDistill}~\cite{shi2021fed}, which
share predictions rather than parameters, the first by combining client models and the
second by averaging label-wise logits; \textbf{FedDF}~\cite{lin2020ensemble},
\textbf{FedGen}~\cite{zhu2021data} and \textbf{FedFTG}~\cite{zhang2022fine}, which all
distill on the server, respectively from the ensemble of client models on unlabeled
data, from a generator that carries the global distribution, and from a generator that
fine-tunes the aggregated model before it is broadcast; and
\textbf{FedACK}~\cite{yang2023fedack}, the GAN-based framework closest to ours in
construction.

Our clients hold subgraphs rather than independent samples, so we add the two federated
\emph{graph} learning methods closest to this setting.
\textbf{GCFL+}~\cite{xie2021gcfl} shares parameters only among clients whose
gradient-norm sequences behave alike; \textbf{FED-PUB}~\cite{baek2023fedpub} probes
client models with a common random graph and turns the similarities among their outputs
into a personalized aggregate and a sparse local mask. Both carry thresholds calibrated
to the gradient and embedding scales of their original benchmarks, and transplanted
unchanged both collapse to FedAvg: GCFL+ never splits its initial cluster and FED-PUB's
similarity matrix becomes uniform. We therefore made GCFL+'s thresholds relative to the
observed first-round gradient norm and centered FED-PUB's functional embeddings before
taking cosine similarity, which leaves the published mechanisms intact while making them
invariant to scale.

Recent centralized detectors (ETS-MM~\cite{etsmm2024}, SEBot~\cite{yang2024sebot},
LMBot~\cite{cai2024lmbot}, BotTrans~\cite{bottrans2023}, CACL~\cite{cacl2024},
BotMoE~\cite{liu2023botmoe}, BotDGT~\cite{he2024dynamicity}) assume the complete graph
and all features. Running them here would require either pooling the clients' data,
which is the assumption the setting exists to remove, or feeding them the fragments
quantified in Section~\ref{setup:graph}, which would report the effect of fragmentation
as a property of those methods. We compare against their published numbers contextually
in Section~\ref{sec:discussion}.

\vspace*{-2pt}
\subsubsection{Evaluation protocol}
\label{setup:protocol}

\mypara{Validation-selected checkpoints.} Each run is reported at the round with the
highest validation accuracy, not at its best test round, since taking a maximum over
$100$ rounds of a quantity no deployment can observe rewards whichever method
oscillates most, and the methods here differ in stability by an order of magnitude.

\mypara{Twenty seeds and paired tests.} Every configuration is repeated over the same
$20$ seeds, each redrawing the partition and reinitializing every model, so any two
methods can be compared seed by seed. We report mean and standard deviation, and every
comparison is a paired $t$-test over the shared seeds; differences that do not reach
$p<0.05$ are reported as ties.

\mypara{Two evaluation targets.} The \emph{global} protocol evaluates the server model on
the pooled test nodes; the \emph{personalized} protocol evaluates each client's own
retained model on its own held-out nodes, averaged over clients holding at least $20$ of
them. The two disagree (Section~\ref{exp:global}), and reporting only one would misrepresent
either the personalized methods or the consensus ones.

\mypara{Where the validation split lives.} A $10\%$ validation split is held by the
server. For every method, it is used to select the reported checkpoint; in \FedRio, it
additionally supplies the controller's reward and the stopping signal for the aggregation
masks. We assess whether these two additional uses affect the comparison through two
controls. The first removes both: it replaces the controller with a random schedule and
bases mask stopping on a fresh batch from the server generator, leaving validation labels
only for model selection. The second assigns \FedRio's internal signals and model
selection to disjoint, stratified halves of the validation split. Under both controls,
\FedRio remains significantly ahead of the strongest baseline in the same six
dataset--skew settings as in Tables~\ref{tab:global-mgtab}
and~\ref{tab:global-twibot} (Table~\ref{tab:validation-controls} in the appendix).
All methods nevertheless require a small labeled server sample for model selection.

\mypara{Backbone.} \FedRio runs with its own cooperative encoder and the baselines with a
two-layer GIN of matched width. The encoder is the part of the design that acts on the
recorded relations, so holding it fixed across methods would remove what is being
evaluated; to keep its share visible, Section~\ref{sec:module-validation} reports
\FedRio on the baselines' encoder.

\vspace*{-2pt}
\subsubsection{Implementation details}
All methods use $10$ clients, $E=5$ local epochs per round, $T=100$ rounds, the Adam
optimizer, batch size $64$, and a two-layer encoder. Baselines use hidden width $64$,
learning rate $10^{-3}$ and weight decay $10^{-4}$. \FedRio's per-cell settings are in
Table~S1 in the appendix; its remaining hyperparameters are fixed across all cells,
with distillation weight $\alpha_{dis}=0.1$, adversarial weight $\gamma_{adv}=0.1$,
contrastive weight $\mu_{con}=0.5$ at temperature $\tau_{con}=0.5$, Gumbel temperature
$\tau=0.5$, a controller learning rate of $5\times10^{-4}$ with discount $\gamma=0.99$
and a $10$-round warm-up, and a server generator of three fully connected layers with
batch normalization and ReLU activations.

%% file: 6-experiment.tex

\vspace*{-3pt}
\subsection{Detection Quality under Label Skew (Q1)}
\label{exp:global}

\begin{table*}[t]
\caption{Global-model accuracy and macro-F1 on MGTAB (\%), mean$\pm$sd over 20 seeds at validation-selected checkpoints. }
\label{tab:global-mgtab}
\centering
\resizebox{0.95\textwidth}{!}{
\begin{tabular}{c|ccccc|ccccc}
\toprule
Metric & \multicolumn{5}{c|}{Accuracy} & \multicolumn{5}{c}{Macro-F1} \\
\midrule
Setting & $\alpha=1.0$ & $\alpha=0.5$ & $\alpha=0.1$ & $\alpha=0.05$ & Avg. rank & $\alpha=1.0$ & $\alpha=0.5$ & $\alpha=0.1$ & $\alpha=0.05$ & Avg. rank \\
\midrule
FedAvg & 85.6$\pm$2.7 & 83.0$\pm$2.7 & 79.5$\pm$2.6 & 78.2$\pm$1.8 & 6.13 & 78.7$\pm$6.1 & 73.8$\pm$6.2 & 66.0$\pm$6.6 & 62.9$\pm$4.9 & 6.00 \\
FedProx & 85.7$\pm$2.5 & 84.0$\pm$2.7 & 81.0$\pm$2.5 & 79.7$\pm$1.3 & 4.13 & 78.9$\pm$5.6 & 75.5$\pm$6.3 & 69.2$\pm$5.8 & 66.6$\pm$3.6 & 4.50 \\
FedDF & 84.4$\pm$3.5 & 79.4$\pm$2.5 & 78.0$\pm$2.8 & 76.0$\pm$1.9 & 9.50 & 76.3$\pm$8.0 & 64.5$\pm$7.3 & 61.2$\pm$9.0 & 53.8$\pm$7.7 & 9.50 \\
FedEnsemble & 88.2$\pm$1.3 & 84.2$\pm$3.5 & 79.7$\pm$4.1 & 75.3$\pm$2.0 & 5.13 & 84.2$\pm$2.1 & 76.5$\pm$8.8 & 66.9$\pm$12.8 & 52.9$\pm$10.7 & 5.00 \\
FedDistill & 85.3$\pm$2.4 & 82.0$\pm$2.2 & 79.4$\pm$2.0 & 78.0$\pm$1.1 & 8.13 & 78.0$\pm$5.3 & 71.3$\pm$5.2 & 65.3$\pm$5.4 & 61.4$\pm$3.5 & 8.00 \\
FedGen & 85.8$\pm$2.3 & 82.8$\pm$2.8 & 79.5$\pm$2.4 & 78.0$\pm$1.4 & 6.00 & 79.2$\pm$5.0 & 73.3$\pm$6.0 & 65.9$\pm$6.0 & 62.6$\pm$4.8 & 6.00 \\
FedFTG & 84.2$\pm$4.2 & 78.9$\pm$2.9 & 77.2$\pm$2.6 & 75.3$\pm$1.7 & 10.75 & 74.9$\pm$10.1 & 62.8$\pm$8.6 & 59.2$\pm$9.8 & 51.0$\pm$6.9 & 10.75 \\
GCFL+ & 85.7$\pm$2.4 & 82.7$\pm$2.8 & 80.6$\pm$2.2 & 79.3$\pm$1.3 & 5.38 & 79.1$\pm$5.3 & 73.1$\pm$6.0 & 69.4$\pm$5.9 & 65.9$\pm$4.6 & 5.00 \\
FED-PUB & 86.2$\pm$3.2 & 82.1$\pm$3.6 & 77.2$\pm$2.5 & 76.5$\pm$1.8 & 7.63 & 79.6$\pm$7.3 & 70.8$\pm$8.9 & 58.2$\pm$8.7 & 55.9$\pm$7.5 & 8.00 \\
\FedACK & 86.4$\pm$1.8 & 84.9$\pm$2.3 & 82.2$\pm$2.1 & 81.2$\pm$1.6 & 2.25 & 80.8$\pm$3.8 & 77.7$\pm$5.0 & 73.1$\pm$4.9 & 70.9$\pm$3.9 & 2.25 \\
\midrule
\FedRio & \textbf{88.9$\pm$0.7} & \textbf{87.8$\pm$1.3} & \textbf{85.2$\pm$1.6} & \textbf{83.7$\pm$1.4} & \textbf{1.00} & \textbf{85.6$\pm$1.0} & \textbf{84.2$\pm$1.9} & \textbf{80.4$\pm$2.9} & \textbf{78.3$\pm$2.6} & \textbf{1.00} \\
\midrule
Gain & \textbf{$\uparrow$\ 0.7$^{\dagger}\sim$4.7} & \textbf{$\uparrow$\ 2.9$^{\dagger}\sim$8.9} & \textbf{$\uparrow$\ 3.0$^{\dagger}\sim$8.0} & \textbf{$\uparrow$\ 2.5$^{\dagger}\sim$8.4} & -- & \textbf{$\uparrow$\ 1.4$^{\dagger}\sim$10.7} & \textbf{$\uparrow$\ 6.5$^{\dagger}\sim$21.4} & \textbf{$\uparrow$\ 7.3$^{\dagger}\sim$22.2} & \textbf{$\uparrow$\ 7.4$^{\dagger}\sim$27.3} & -- \\
\bottomrule
\end{tabular}}
\vspace{-3pt}
\end{table*}

\begin{table*}[t]
\caption{Global-model accuracy and macro-F1 on TwiBot-20 (\%), mean$\pm$sd over 20 seeds at validation-selected checkpoints. Best per column in bold. ``Avg. rank'' is the mean rank among the eleven methods over the four partitions, tied values sharing the average of the ranks they span. ``Gain'' is \FedRio\ minus the strongest baseline of that column, with $^{\dagger}$ marking $p<0.05$ under a one-sided paired $t$-test over the shared seeds.}
\label{tab:global-twibot}
\centering
\resizebox{0.95\textwidth}{!}{
\begin{tabular}{c|ccccc|ccccc}
\toprule
Metric & \multicolumn{5}{c|}{Accuracy} & \multicolumn{5}{c}{Macro-F1} \\
\midrule
Setting & $\alpha=1.0$ & $\alpha=0.5$ & $\alpha=0.1$ & $\alpha=0.05$ & Avg. rank & $\alpha=1.0$ & $\alpha=0.5$ & $\alpha=0.1$ & $\alpha=0.05$ & Avg. rank \\
\midrule
FedAvg & \textbf{85.7$\pm$0.8} & 84.3$\pm$1.5 & 80.5$\pm$4.0 & 78.4$\pm$2.2 & 5.38 & \textbf{85.4$\pm$0.7} & 84.0$\pm$1.4 & 80.1$\pm$4.0 & 77.4$\pm$2.7 & 5.25 \\
FedProx & 85.5$\pm$0.8 & 84.3$\pm$1.4 & 80.4$\pm$4.0 & 78.3$\pm$2.2 & 7.63 & 85.2$\pm$0.8 & 84.1$\pm$1.3 & 80.0$\pm$4.0 & 77.3$\pm$2.7 & 7.13 \\
FedDF & \textbf{85.7$\pm$0.7} & 84.3$\pm$1.4 & 79.9$\pm$4.1 & 78.5$\pm$3.2 & 6.13 & \textbf{85.4$\pm$0.7} & 84.1$\pm$1.3 & 79.6$\pm$4.1 & 77.1$\pm$4.3 & 6.50 \\
FedEnsemble & \textbf{85.7$\pm$0.7} & \textbf{85.1$\pm$1.0} & 80.1$\pm$7.1 & 69.3$\pm$15.1 & 5.88 & 85.3$\pm$0.7 & 84.5$\pm$1.0 & 79.3$\pm$8.2 & 64.3$\pm$19.9 & 7.50 \\
FedDistill & 85.5$\pm$0.7 & 84.7$\pm$1.1 & 81.2$\pm$3.5 & 78.9$\pm$2.5 & 4.75 & \textbf{85.4$\pm$0.7} & 84.4$\pm$1.1 & 80.8$\pm$3.5 & 77.9$\pm$3.0 & 3.00 \\
FedGen & \textbf{85.7$\pm$0.8} & 84.4$\pm$1.2 & 80.2$\pm$4.0 & 78.5$\pm$2.3 & 5.00 & \textbf{85.4$\pm$0.8} & 84.2$\pm$1.2 & 79.8$\pm$4.0 & 77.5$\pm$2.8 & 4.88 \\
FedFTG & 85.6$\pm$0.7 & 84.4$\pm$1.5 & 80.5$\pm$3.8 & 76.8$\pm$5.9 & 6.50 & 85.3$\pm$0.7 & 84.2$\pm$1.5 & 80.0$\pm$3.8 & 74.4$\pm$9.6 & 6.75 \\
GCFL+ & 85.6$\pm$0.8 & 83.6$\pm$1.8 & 79.8$\pm$4.1 & 78.1$\pm$3.0 & 8.75 & \textbf{85.4$\pm$0.8} & 83.4$\pm$1.8 & 79.5$\pm$4.0 & 77.1$\pm$3.0 & 7.38 \\
FED-PUB & 85.6$\pm$0.9 & 84.7$\pm$1.5 & 82.2$\pm$4.2 & 79.9$\pm$5.2 & 3.63 & 85.2$\pm$0.9 & 84.3$\pm$1.5 & 81.7$\pm$4.1 & 78.2$\pm$7.1 & 4.38 \\
\FedACK & 84.5$\pm$1.5 & 81.0$\pm$4.0 & 76.2$\pm$8.8 & 75.6$\pm$6.6 & 10.75 & 84.2$\pm$1.4 & 80.8$\pm$4.0 & 75.0$\pm$10.8 & 74.0$\pm$7.4 & 10.75 \\
\midrule
\FedRio & \textbf{85.7$\pm$1.1} & \textbf{85.1$\pm$1.2} & \textbf{84.6$\pm$1.4} & \textbf{83.9$\pm$1.7} & \textbf{1.63} & 85.3$\pm$1.0 & \textbf{84.6$\pm$1.2} & \textbf{83.9$\pm$1.6} & \textbf{83.3$\pm$1.8} & \textbf{2.50} \\
\midrule
Gain & \textbf{$\uparrow$\ 0.0$^{\dagger}\sim$1.2}  & \textbf{$\uparrow$\ 0.0$^{\dagger}\sim$4.1} & \textbf{$\uparrow$\ 2.4$^{\dagger}\sim$8.4} & \textbf{$\uparrow$\ 4.0$^{\dagger}\sim$14.6} & -- & $-0.1\sim1.1$ & \textbf{$\uparrow$\ 0.1$^{\dagger}\sim$3.8} & \textbf{$\uparrow$\ 2.2$^{\dagger}\sim$8.9} & \textbf{$\uparrow$\ 5.1$^{\dagger}\sim$19.0} &-- \\
\bottomrule
\end{tabular}}
\end{table*}

Tables~\ref{tab:global-twibot} and~\ref{tab:global-mgtab} report the accuracy and
macro-F1 of the \emph{global} model. Every number is the mean over the same 20 seeds at
the checkpoint selected by validation accuracy, with identical splits, partitions and
local objectives across methods; a cell is marked with $\dagger$ only when \FedRio\
beats the \emph{strongest} baseline of that column under a one-sided paired $t$-test at
$p<0.05$, and unmarked cells are not claimed as improvements. 

On MGTAB \FedRio\ is significantly ahead at every skew level, and by far more on
macro-F1 ($1.4$ to $7.4$ points) than on accuracy ($0.7$ to $3.0$). That gap is the
substance rather than a detail: humans outnumber bots roughly three to one here, so a
model drifting toward the majority class keeps a respectable accuracy while its
macro-F1 collapses. At $\alpha=0.05$ FedAvg holds $78.2\%$ accuracy with $62.9$
macro-F1 and FED-PUB $76.5\%$ with $55.9$, against $83.7\%$ and $78.3$ for
\FedRio. On TwiBot-20 the advantage is compressed under mild skew and separates as the
partition sharpens: at $\alpha=1$ nine of the ten baselines lie within $0.2$ macro-F1
points of one another and \FedRio\ sits inside that band, a tenth of a point behind
FedDistill, and at $\alpha=0.5$ it is level with FedEnsemble; at
$\alpha=0.1$ and $\alpha=0.05$ it leads the strongest baseline by $2.2$ and $5.1$
points, both significant.

From $\alpha=1$ to $\alpha=0.05$, FedAvg loses $8.0$ macro-F1 points on TwiBot-20 and $15.8$ on MGTAB
while \FedRio\ loses $2.0$ and $7.3$, so the margin grows monotonically with skew on
both benchmarks. The average ranks say the
same without reference to any single partition: \FedRio\ holds rank $1.00$ on MGTAB
under both measures and the best rank on TwiBot-20 as well, $1.63$ on accuracy and
$2.50$ on macro-F1 against FedDistill's $3.00$. The two remain closest where clients are
alike, since FedDistill is strongest under the mildest partition and \FedRio\ under the
sharpest. Stability moves the same way: at $\alpha=0.05$ \FedRio\ varies by $1.8$
and $2.6$ macro-F1 points across seeds against $2.7$ to $19.9$ for the baselines.

\begin{table*}[t]
\caption{Ablation on the tuned operating points (macro-F1 \%, 20 seeds, validation-selected checkpoints). $^{*}$ marks a variant significantly below the full model at $p<0.05$ under a one-sided paired $t$-test over shared seeds; unmarked rows are within seed noise.}
\label{tab:module-ablation}
\centering
\resizebox{0.85\textwidth}{!}{
\begin{tabular}{c|cccc|cccc}
\toprule
Dataset & \multicolumn{4}{c|}{TwiBot-20} & \multicolumn{4}{c}{MGTAB} \\
\midrule
Setting & $\alpha=1.0$ & $\alpha=0.5$ & $\alpha=0.1$ & $\alpha=0.05$ & $\alpha=1.0$ & $\alpha=0.5$ & $\alpha=0.1$ & $\alpha=0.05$ \\
\midrule
\FedRio-NA & 85.5$\pm$0.8 & 83.2$\pm$3.1$^{*}$ & 84.0$\pm$1.2 & 82.5$\pm$1.8$^{*}$ & 84.9$\pm$1.7 & 81.7$\pm$2.9$^{*}$ & 79.0$\pm$3.4$^{*}$ & 77.3$\pm$3.7 \\
\FedRio-NR & 85.0$\pm$0.8$^{*}$ & 82.1$\pm$2.5$^{*}$ & 83.6$\pm$2.2 & 80.5$\pm$2.6$^{*}$ & 85.7$\pm$1.2 & 83.8$\pm$2.4 & 79.7$\pm$2.5 & 78.5$\pm$3.0 \\
\FedRio-NC & 84.8$\pm$1.0 & 83.6$\pm$1.6$^{*}$ & 82.7$\pm$2.1$^{*}$ & 81.5$\pm$2.3$^{*}$ & 85.5$\pm$1.5 & 83.7$\pm$1.8 & 78.6$\pm$3.3$^{*}$ & 76.0$\pm$3.8$^{*}$ \\
\FedRio-NA\&NR & 85.2$\pm$0.7 & 81.5$\pm$3.1$^{*}$ & 82.3$\pm$4.3$^{*}$ & 75.9$\pm$7.3$^{*}$ & 84.6$\pm$2.1$^{*}$ & 81.6$\pm$2.4$^{*}$ & 80.4$\pm$1.8 & 78.0$\pm$2.1 \\
\midrule
\FedRio & 85.3$\pm$1.0 & 84.6$\pm$1.2 & 83.9$\pm$1.6 & 83.3$\pm$1.8 & 85.6$\pm$1.0 & 84.2$\pm$1.9 & 80.4$\pm$2.9 & 78.3$\pm$2.6 \\
\bottomrule
\end{tabular}}
\end{table*}

\mypara{How much of this is the encoder.}
Since the backbones differ, part of the margin belongs to the cooperative encoder
rather than to the federated algorithm. The \FedRio-NC row of
Table~\ref{tab:module-ablation} gives \FedRio\ the baselines' GIN encoder: that variant
still leads the strongest baseline by $5.1$ and $5.5$ macro-F1 points on MGTAB at
$\alpha=0.05$ and $\alpha=0.1$ and by $3.3$ on TwiBot-20 at $\alpha=0.05$, while under
mild skew it falls marginally below the best baseline on TwiBot-20. The encoder is
therefore worth one to two points throughout and is not the source of the advantage
where the advantage is large.

\begin{table*}[t]
\caption{Personalized evaluation (\%), mean$\pm$sd over 20 seeds: each client is scored on its own held-out split with its own retained model, at the validation-selected round. Clients holding fewer than 20 test nodes are excluded. ``Worst-client'' is the lowest per-client macro-F1 in a run. Best per column in bold; $^{\dagger}$ marks $p<0.05$ against the strongest baseline.}
\label{tab:personal}
\centering
\resizebox{0.95\textwidth}{!}{
\begin{tabular}{c|ccccc|ccccc}
\toprule
Dataset & \multicolumn{5}{c|}{TwiBot-20} & \multicolumn{5}{c}{MGTAB} \\
\midrule
Setting & $\alpha=1.0$ & $\alpha=0.5$ & $\alpha=0.1$ & $\alpha=0.05$ & Avg. rank & $\alpha=1.0$ & $\alpha=0.5$ & $\alpha=0.1$ & $\alpha=0.05$ & Avg. rank \\
\midrule
\multicolumn{11}{c}{\emph{Personalized macro-F1, averaged over clients}} \\
\midrule
FedAvg & 79.8$\pm$4.6 & 78.0$\pm$6.0 & 74.7$\pm$5.5 & 77.8$\pm$11.4 & 4.38 & 75.8$\pm$4.0 & 68.5$\pm$5.2 & 65.2$\pm$7.2 & 64.9$\pm$10.9 & 6.00 \\
FedProx & 79.5$\pm$3.9 & 77.2$\pm$7.0 & 73.6$\pm$5.6 & 77.5$\pm$12.8 & 7.00 & 74.6$\pm$2.8 & 67.8$\pm$3.9 & 64.6$\pm$7.5 & 60.2$\pm$9.3 & 9.50 \\
FedDF & 79.4$\pm$4.6 & 77.2$\pm$6.6 & 73.4$\pm$6.0 & 77.3$\pm$10.6 & 7.88 & 74.2$\pm$3.9 & 69.3$\pm$5.6 & 65.9$\pm$9.1 & 71.9$\pm$7.5 & 5.50 \\
FedEnsemble & \textbf{80.2$\pm$5.0} & \textbf{78.1$\pm$6.2} & 74.7$\pm$7.8 & 80.1$\pm$12.1 & \textbf{2.63} & 77.1$\pm$3.3 & 71.2$\pm$5.4 & 71.9$\pm$11.9 & 70.6$\pm$15.0 & 3.00 \\
FedDistill & 80.0$\pm$3.9 & 77.8$\pm$5.9 & 76.3$\pm$7.1 & 78.9$\pm$8.6 & \textbf{2.63} & 75.5$\pm$3.9 & 69.2$\pm$5.3 & 65.8$\pm$7.8 & 64.8$\pm$10.8 & 6.25 \\
FedGen & 79.4$\pm$4.3 & 76.8$\pm$6.0 & 75.8$\pm$5.6 & 78.5$\pm$10.0 & 6.38 & 75.6$\pm$4.4 & 68.6$\pm$4.6 & 64.2$\pm$8.4 & 63.7$\pm$9.7 & 7.38 \\
FedFTG & 78.9$\pm$4.1 & 77.8$\pm$6.5 & 73.7$\pm$5.4 & 77.0$\pm$11.2 & 7.88 & 74.7$\pm$4.2 & 69.4$\pm$6.2 & 66.5$\pm$9.9 & 71.6$\pm$11.4 & 4.75 \\
GCFL+ & 79.1$\pm$4.5 & 76.2$\pm$5.9 & 75.5$\pm$8.0 & 78.7$\pm$10.4 & 7.00 & 75.6$\pm$4.2 & 68.1$\pm$4.3 & 64.7$\pm$8.4 & 61.4$\pm$8.5 & 7.88 \\
FED-PUB & 79.7$\pm$4.6 & 77.2$\pm$6.2 & 76.2$\pm$9.1 & 76.7$\pm$10.9 & 6.13 & 80.7$\pm$5.3 & 74.1$\pm$6.3 & \textbf{75.6$\pm$9.4} & \textbf{79.6$\pm$12.3} & \textbf{1.50} \\
\FedACK & 78.5$\pm$5.1 & 75.2$\pm$7.2 & 72.8$\pm$7.9 & 77.2$\pm$8.7 & 10.50 & 73.8$\pm$3.8 & 66.7$\pm$4.2 & 64.1$\pm$7.0 & 59.3$\pm$6.9 & 11.00 \\
\midrule
\FedRio & 79.7$\pm$4.3 & 77.0$\pm$6.6 & \textbf{78.3$\pm$8.2} & \textbf{80.9$\pm$8.0} & 3.63 & \textbf{83.3$\pm$4.7$^{\dagger}$} & \textbf{79.2$\pm$3.5$^{\dagger}$} & 67.9$\pm$10.5 & 62.5$\pm$12.0 & 3.25 \\
\midrule
\multicolumn{11}{c}{\emph{Worst-client macro-F1}} \\
\midrule
FedAvg & 61.2$\pm$10.3 & 53.7$\pm$10.2 & 49.6$\pm$1.3 & 54.9$\pm$13.3 & 3.88 & 50.4$\pm$4.4 & 48.9$\pm$1.6 & 48.6$\pm$1.2 & 48.3$\pm$2.8 & 8.13 \\
FedProx & 60.7$\pm$10.6 & 53.2$\pm$10.2 & 49.2$\pm$0.9 & 53.6$\pm$14.7 & 6.75 & 49.3$\pm$0.4 & 49.1$\pm$0.5 & 48.7$\pm$1.0 & 49.0$\pm$0.9 & 7.63 \\
FedDF & 60.0$\pm$10.8 & 53.7$\pm$10.0 & 49.3$\pm$0.8 & 53.9$\pm$15.1 & 6.50 & 49.3$\pm$0.6 & 49.3$\pm$1.5 & 48.4$\pm$1.5 & 50.0$\pm$3.0 & 7.25 \\
FedEnsemble & \textbf{62.7$\pm$11.2} & 55.3$\pm$10.7 & 48.4$\pm$4.1 & \textbf{60.4$\pm$19.1} & 3.25 & 49.7$\pm$0.3 & 49.1$\pm$1.2 & \textbf{49.3$\pm$3.7} & 53.3$\pm$13.1 & 4.38 \\
FedDistill & 58.9$\pm$10.4 & \textbf{55.6$\pm$10.8} & 49.9$\pm$3.9 & 54.4$\pm$13.1 & 4.38 & 50.2$\pm$3.7 & 49.3$\pm$0.5 & 48.7$\pm$1.3 & 50.1$\pm$4.4 & 4.50 \\
FedGen & 60.0$\pm$10.8 & 51.3$\pm$7.8 & 49.5$\pm$0.6 & 55.0$\pm$13.5 & 6.13 & 50.0$\pm$2.6 & 49.0$\pm$1.9 & 48.5$\pm$1.5 & 48.8$\pm$1.4 & 8.63 \\
FedFTG & 58.6$\pm$9.8 & 54.1$\pm$10.4 & 48.9$\pm$1.5 & 54.0$\pm$13.2 & 7.25 & 49.5$\pm$0.3 & 49.6$\pm$3.5 & 47.6$\pm$3.0 & 50.8$\pm$6.7 & 6.00 \\
GCFL+ & 60.3$\pm$10.1 & 53.4$\pm$9.9 & 48.2$\pm$8.5 & 54.7$\pm$13.0 & 7.00 & 50.1$\pm$3.4 & 49.1$\pm$1.1 & 48.8$\pm$1.1 & 48.8$\pm$1.4 & 6.50 \\
FED-PUB & 58.6$\pm$10.6 & 52.0$\pm$10.3 & 48.9$\pm$13.5 & 52.7$\pm$15.6 & 9.50 & 55.6$\pm$11.4 & 49.4$\pm$0.8 & 48.9$\pm$1.4 & \textbf{57.5$\pm$16.6} & \textbf{2.13} \\
\FedACK & 58.9$\pm$10.2 & 53.9$\pm$10.4 & 47.9$\pm$8.1 & 51.6$\pm$14.3 & 8.88 & 49.1$\pm$0.6 & 49.1$\pm$0.6 & 48.6$\pm$1.2 & 49.1$\pm$0.6 & 8.25 \\
\midrule
\FedRio & 60.6$\pm$10.9 & 54.2$\pm$10.3 & \textbf{52.7$\pm$11.1} & 56.8$\pm$15.1 & \textbf{2.50} & \textbf{56.8$\pm$11.6} & \textbf{51.4$\pm$4.7$^{\dagger}$} & 48.9$\pm$3.0 & 49.6$\pm$4.3 & 2.63 \\
\bottomrule
\end{tabular}}
\vspace{-3pt}
\end{table*}

\mypara{Personalized detection quality.}
The tables above score one server model on the pooled test set. A platform joining the
federation experiences something different: its own model on its own accounts.
Table~\ref{tab:personal} reports that read-out, and it does not reproduce the global
ordering. \FedRio\ leads on TwiBot-20 at the two sharpest partitions and on MGTAB at the
two mildest, where it is $2.6$ and $5.1$ points ahead of FED-PUB, but on MGTAB at
$\alpha=0.1$ and $\alpha=0.05$ it is behind FED-PUB by $7.7$ and $17.1$ points, and its
average rank on TwiBot-20 ($3.63$) trails FedEnsemble's and FedDistill's ($2.63$).

The two protocols reward different things, and FED-PUB is the clearest case. It
aggregates by client similarity and keeps a sparse local mask, which specializes each
client to its own slice; on the global MGTAB tables it is among the weakest of the
eleven methods at exactly those two partitions ($58.2$ and $55.9$ macro-F1 against
$80.4$ and $78.3$ for \FedRio). When a partition leaves a client holding almost a single
class, that specialization is what the client's own held-out split rewards and what the
pooled split penalizes, so the disagreement between the two tables is a trade between
objectives rather than a contradiction, and it appears where the objectives diverge
most. In the lower block of the same table, which scores the worst-served client in each run,
the ordering moves back: \FedRio\ holds the best average rank on TwiBot-20 ($2.50$
against FedEnsemble's $3.25$) and the second best on MGTAB ($2.63$ behind FED-PUB's
$2.13$).

\begin{table}[t]
\caption{A client-specific broadcast: the global term of Eq.~\ref{fedrio:eq:updateclient} is a similarity-weighted mixture of the uploaded models instead of the single consensus model. Personalized macro-F1 averaged over clients, $20$ seeds; the consensus column repeats \FedRio\ from Table~\ref{tab:personal} and $\Delta$ is the change against it. ``$\dagger$'' marks a paired difference from FED-PUB significant at $p<0.05$.}
\label{tab:persagg}
\centering
\scalebox{0.9}{
\begin{tabular}{c|c|ccc|c}
\toprule
Dataset & $\alpha$ & Consensus & Mixture & $\Delta$ & FED-PUB \\
\midrule
TwiBot-20 & 1.0 & 79.7 & 80.3 & $+0.6$ & 79.7 \\
 & 0.5 & 77.0 & 78.2 & $+1.2$ & 77.2 \\
 & 0.1 & 78.3 & 79.5 & $+1.2$ & 76.2 \\
 & 0.05 & 80.9 & 79.7 & $-1.2$ & 76.7 \\
\midrule
MGTAB & 1.0 & 83.3 & \textbf{83.6}$^{\dagger}$ & $+0.3$ & 80.7 \\
 & 0.5 & 79.2 & \textbf{80.2}$^{\dagger}$ & $+1.0$ & 74.1 \\
 & 0.1 & 67.9 & 68.0$^{\dagger}$ & $+0.1$ & 75.6 \\
 & 0.05 & 62.5 & 66.7$^{\dagger}$ & $+4.2$ & 79.6 \\
\bottomrule
\end{tabular}}
\end{table}

Since this read-out is the one \FedRio\ was not built for, we evaluate a variant of the
client update of Eq.~\ref{fedrio:eq:updateclient} in which the global term
$\Theta_{t-1}$ is client-specific: the server sends client $k$ a mixture
$\sum_j s_{kj}\Theta^j_t$ of the uploaded models, weighted by the similarity $s_{kj}$
between clients, rather than the single consensus model, with the mixing weight chosen
per cell on the personalized validation split. This variant is used nowhere else in the
paper. Table~\ref{tab:persagg} reports it. The mixture improves on the consensus model
of Table~\ref{tab:personal} in seven of the eight cells, the exception being TwiBot-20
at $\alpha=0.05$, where it costs $1.2$ points. It puts \FedRio\ significantly ahead of
FED-PUB on MGTAB under mild and moderate skew ($83.6$ against $80.7$, and $80.2$ against
$74.1$), but it does not recover the two cells at issue: there it gains $0.1$ and $4.2$
points and remains $7.6$ and $12.9$ behind. The reason is structural. Mixing by
similarity pools clients whose label distributions already agree, so under a partition
that leaves each client with almost one class it reinforces the very bias the client
would need to correct.

We therefore claim the global result and not a general personalization advantage.
\FedRio\ is the stronger choice when the federation is judged by the model it produces
jointly; a deployment in which each participant is judged only on its own accounts under
extreme skew is better served by a personalization-first method such as FED-PUB.

\vspace*{-3pt}
\subsection{Efficiency (Q2)}
\label{exp:efficiency}

\begin{table}[t]
\caption{Cost of reaching a usable model. ``Rounds'' is the first round at which validation accuracy reaches $80\%$, averaged over the seeds that reach it; a dash means it was never reached within $100$ rounds, and the figure in parentheses is the total parameters transmitted over those rounds. ``Payload'' is the parameters a client uploads per round, both in millions, given as a range where \FedRio's tuned encoder widths differ; ``Time'' is wall-clock seconds per round on one H200. The last row runs \FedRio\ at the $64$ units every baseline uses.}
\label{tab:efficiency}
\centering
\resizebox{\columnwidth}{!}{
\begin{tabular}{c|cc|cc|cc}
\toprule
& \multicolumn{2}{c|}{TwiBot-20} & \multicolumn{2}{c|}{MGTAB} & & \\
Method & $\alpha=0.5$ & $0.05$ & $0.5$ & $0.05$ & Payload (M) & Time (s) \\
\midrule
FedAvg & 27.2 (21.9) & 66.5 (53.5) & 23.8 (19.0) & 23.2 (18.6) & 0.80 & 0.3 \\
FedProx & 22.8 (18.4) & 66.5 (53.5) & 9.6 (7.7) & 10.8 (8.7) & 0.80 & 0.3 \\
FedDF & 28.2 (22.7) & 55.7 (44.8) & 33.0 (26.4) & -- & 0.80 & 0.4 \\
FedEnsemble & 12.2 (9.8) & 64.0 (51.5) & 14.0 (11.2) & -- & 0.80 & 0.3 \\
FedDistill & 22.4 (18.0) & 54.7 (44.0) & 26.8 (21.5) & 25.8 (20.6) & 0.80 & 0.3 \\
FedGen & 23.8 (19.2) & 48.0 (38.6) & 22.0 (17.6) & 26.0 (20.8) & 0.80 & 0.4 \\
FedFTG & 30.0 (24.2) & 76.0 (61.2) & 32.0 (25.6) & -- & 0.80 & 0.5 \\
GCFL+ & 26.2 (21.1) & 69.5 (56.0) & 24.6 (19.7) & 31.8 (25.5) & 0.80 & 0.3 \\
FED-PUB & 22.4 (18.0) & 53.0 (42.7) & 13.3 (10.7) & 11.5 (9.2) & 0.80 & 0.3 \\
\FedACK & 37.2 (30.0) & 42.0 (33.8) & 7.0 (5.6) & 8.8 (7.1) & 0.80 & 1.6 \\
\midrule
\FedRio & 17.2 (60.5) & 21.4 (24.5) & 2.4 (8.4) & 3.4 (3.9) & 1.14--3.52 & 1.7 \\
\FedRio, $64$ units & 31.0 (35.3) & 23.2 (26.5) & 4.0 (4.6) & 3.4 (3.9) & 1.14 & 1.4 \\
\bottomrule
\end{tabular}}
\end{table}

Table~\ref{tab:efficiency} reports what a round costs, how many rounds are needed, and
the product of the two. The round count is the first round at which validation accuracy
reaches $80\%$, so it measures time to a usable model rather than time to convergence,
and the figure beside it is the total number of parameters a client transmits before the
federation has such a model. Both the payload and the per-round wall time are measured
from the runs rather than derived from the bounds in the appendix.

\FedRio\ reaches the target in the fewest rounds in three of the four cells, and the
saving is largest exactly where the partition is sharp: $2.4$ and $3.4$ rounds on MGTAB
against $7.0$ to $33.0$ for the baselines, and $21.4$ rounds on TwiBot-20 at
$\alpha=0.05$ against $42.0$ to $76.0$. Several baselines never reach the target at
all under sharp skew, which is the same behavior the accuracy tables show from a
different angle.

The per-round cost runs the other way and we report it plainly. A \FedRio\ client
uploads between $1.14$M and $3.52$M parameters per round against $0.80$M for every
baseline, since the cooperative encoder is larger than the GIN encoder the baselines
share, and a round costs $1.7$ seconds against
$0.3$ for the lightweight methods and $1.6$ for \FedACK. Part of that spread is model
size rather than design, because tuning selected a $128$-unit encoder in half the cells
while every baseline runs $64$ units, so the last row of the table repeats \FedRio\ at
the width the baselines use. Held there its payload is $1.14$M, $1.4$ times a
baseline's rather than up to $4.4$ times, and its round count changes little except on
TwiBot-20 under mild skew. Nor does the accuracy comparison rest on the larger encoder:
the \FedRio-NC row of Table~\ref{tab:module-ablation} runs the baselines' own GIN encoder
and still leads the strongest baseline by $3.3$ to $5.5$ macro-F1 points where the margin
is large, so the extra parameters are worth one to two points.

Multiplying the round count by the payload gives the figure in parentheses, the total a
client transmits before the federation has a usable model. On TwiBot-20 at $\alpha=0.5$
that total is the highest in its column, $35$M at equal width and $60.5$M at the tuned
$128$-unit point against FedEnsemble's $9.8$M. In the remaining three cells \FedRio\ at
equal width reaches the target on fewer transmitted parameters than any baseline.

\vspace*{-3pt}
\subsection{Feature Space Consistency (Q3)}
\label{exp:feature_space_consistency}

\begin{table}[t]
\caption{Representation alignment across clients: linear CKA between the encoders of every pair of clients, and between each client and the server, at the selected checkpoint. The two columns of each block differ only in the contrastive term. ``Seeds'' counts the runs in which CKA is defined; runs where a client collapses to a constant representation are dropped, not imputed.}
\label{tab:cka}
\centering
\scalebox{0.90}{
\begin{tabular}{cc|cc|cc|c}
\toprule
& & \multicolumn{2}{c|}{client--client CKA} & \multicolumn{2}{c|}{client--server CKA} & \\
Dataset & $\alpha$ & with & without & with & without & Seeds \\
\midrule
TwiBot-20 & 1.0 & 0.96 & 0.83 & 0.97 & 0.89 & 5 \\
TwiBot-20 & 0.5 & 0.85 & 0.60 & 0.89 & 0.66 & 5 \\
TwiBot-20 & 0.1 & 0.91 & 0.69 & 0.95 & 0.78 & 5 \\
TwiBot-20 & 0.05 & 0.66 & 0.59 & 0.79 & 0.68 & 5 \\
MGTAB & 1.0 & 0.77 & 0.60 & 0.87 & 0.75 & 10 \\
MGTAB & 0.5 & 0.69 & 0.56 & 0.80 & 0.69 & 10 \\
MGTAB & 0.1 & 0.90 & 0.81 & 0.93 & 0.87 & 10 \\
MGTAB & 0.05 & 0.91 & 0.83 & 0.94 & 0.89 & 10 \\
\bottomrule
\end{tabular}}
\end{table}

Table~\ref{tab:cka} asks whether clients converge on a common representation of accounts.
Every client encoder, restored to its validation-selected round, embeds the same held-out
accounts, and we report linear centered kernel alignment~\cite{kornblith2019similarity} (CKA) between the resulting
representations for every pair of clients and between each client and the server encoder.
CKA compares two encodings of identical inputs by the geometry they induce rather than
coordinate by coordinate: it is invariant to rotation and to isotropic rescaling, so a
client is not penalized for arranging the same information along different axes, and it
runs from $1$, when two encoders agree up to those transformations, to $0$, when they are
unrelated. Setting the contrastive weight to zero is the only difference between the two
columns of each block, so the comparison isolates that term rather than the framework.

The contrastive term raises alignment in all eight cells, by between $0.07$ and $0.25$,
and alignment to the server model moves with it. The largest gains are on TwiBot-20
under moderate skew, where client--client CKA rises from $0.60$ to $0.85$ at
$\alpha=0.5$ and from $0.69$ to $0.91$ at $\alpha=0.1$; on MGTAB, whose denser graph
already aligns clients to some degree, the term adds between $0.08$ and $0.17$. One
qualification belongs here: under the sharpest skew on TwiBot-20 the gap narrows to
$0.07$, so the term does not rescue alignment when a client holds almost one class,
which is also the regime in which a client occasionally collapses to a constant
representation and the measurement becomes undefined. This sweep uses five seeds on
TwiBot-20 and ten on MGTAB rather than the twenty used elsewhere, since each run must
retain all ten client encoders at the selected round to be measurable at all.

\vspace*{-3pt}
\subsection{Robustness and Privacy Noise (Q4)}
\label{exp:sensitivity}

\begin{table}[t]
\caption{Robustness of \FedRio\ (macro-F1 \%, 10 seeds). The upper block varies the number of participating clients on a fixed dataset; the lower clips every uploaded controller state and perturbs it with Gaussian noise of the stated scale. $^{*}$ marks a setting significantly below its column's default at $p<0.05$ under a one-sided paired $t$-test.}
\label{tab:robustness}
\centering
\scalebox{0.98}{
\begin{tabular}{c|cc|cc}
\toprule
& \multicolumn{2}{c|}{TwiBot-20} & \multicolumn{2}{c}{MGTAB} \\
Setting & $\alpha=0.5$ & $\alpha=0.05$ & $\alpha=0.5$ & $\alpha=0.05$ \\
\midrule
\multicolumn{5}{c}{\emph{Number of clients}} \\
\midrule
$K=5$ & 84.9$\pm$1.8 & 83.3$\pm$2.8 & 84.7$\pm$2.0 & 77.5$\pm$3.7 \\
$K=10$ (default) & 84.5$\pm$1.9 & 83.3$\pm$1.4 & 84.5$\pm$1.7 & 78.4$\pm$2.8 \\
$K=20$ & 85.1$\pm$0.9 & 82.8$\pm$1.6 & 84.6$\pm$1.1 & 77.7$\pm$3.0 \\
$K=50$ & 84.4$\pm$0.9 & 81.7$\pm$2.8$^{*}$ & 83.9$\pm$1.2 & 78.6$\pm$3.4 \\
\midrule
\multicolumn{5}{c}{\emph{Gaussian noise on the uploaded controller state}} \\
\midrule
$\sigma=0.0$ (default) & 84.8$\pm$1.5 & 83.0$\pm$1.7 & 84.0$\pm$1.8 & 78.3$\pm$2.0 \\
$\sigma=0.01$ & 84.6$\pm$1.6 & 83.4$\pm$1.7 & 84.0$\pm$1.8 & 76.6$\pm$4.0$^{*}$ \\
$\sigma=0.05$ & 84.5$\pm$1.5 & 83.1$\pm$1.8 & 84.2$\pm$1.7 & 77.5$\pm$4.4 \\
$\sigma=0.1$ & 84.6$\pm$1.5 & 83.3$\pm$1.9 & 84.0$\pm$1.8 & 77.4$\pm$3.7 \\
$\sigma=0.5$ & 84.7$\pm$1.6 & 83.2$\pm$1.4 & 84.1$\pm$1.6 & 78.2$\pm$2.4 \\
\bottomrule
\end{tabular}}
\end{table}

Table~\ref{tab:robustness} varies two quantities a deployment does not get to choose:
how many platforms join, and how much noise is added to what they upload.

Federation size has no consistent effect between $K=5$ and $K=50$. The one cell that
crosses the significance threshold is $K=50$ on TwiBot-20 at $\alpha=0.05$, where each
client is left with roughly $165$ training accounts and the score falls $1.6$ points
below the ten-client default; the same setting on MGTAB, whose graph is twenty times
denser and leaves each client far more structure to exploit, does not move. We read this
as a floor on what a client must hold rather than as a scaling limit of the method.

The noise block is the quantitative version of the privacy statement in
Section~\ref{sec:privacy}. Each client's uploaded controller state is clipped and
perturbed with Gaussian noise before it leaves the client, over a fiftyfold range of
scales. Detection quality does not degrade across that range: the largest deviation from
the noiseless default is $1.7$ macro-F1 points, a single cell crosses the one-sided
threshold---MGTAB at $\alpha=0.05$ and $\sigma=0.01$, the \emph{weakest} perturbation we
apply---and the effect is non-monotone, since at $\sigma=0.5$ two of the four columns sit
above the noiseless default and the other two within $0.1$ points of it. We therefore
report that this state can be perturbed at no measurable cost in this range, which is a
statement about the range tested rather than a privacy guarantee.

\vspace*{-3pt}
\subsection{Module Validation (Q5)}
\label{sec:module-validation}

Table~\ref{tab:module-ablation} removes components from the tuned operating point of
each cell, so every arm is compared against the configuration the main table actually
reports. \FedRio-NA drops the neuron-level aggregation, \FedRio-NR the
reinforcement-learned client controller, \FedRio-NA\&NR both, and \FedRio-NC the
cooperative encoder.

Neither proposed module earns much when clients are alike: at $\alpha=1$ the only single
removal a paired test detects is the controller on TwiBot-20, worth $0.3$ points. That is
what the design predicts, since an update-rate policy and a per-coordinate admission rule
have nothing to arbitrate when clients hold near-identical distributions. Both matter
under moderate skew---$1.4$ and $2.5$ points on TwiBot-20 at $\alpha=0.5$, $2.5$ for the
aggregation on MGTAB---and they interact rather than add, since removing both at once
costs $7.4$ points on TwiBot-20 at $\alpha=0.05$, twice the sum of the two. Where the
margins are largest they explain least: on MGTAB at $\alpha=0.05$ neither is detectable
while the cooperative encoder costs $2.3$ points, and at $\alpha=0.1$ its $1.8$ exceeds
the aggregation's $1.4$. We report that rather than the attribution we expected.

We analysed why the encoder matters unevenly. The fraction of a client's edges
surviving a layer is measured rather than set, and averages $27\%$ to $49\%$ on MGTAB
against $1\%$ to $11\%$ on TwiBot-20, whose follow graph is sparser (average degree $7.8$
against $177$) and less homophilous ($0.59$ against $0.75$). On TwiBot-20 the module
learns to propagate almost nothing and to classify accounts on their attributes alone,
which is worth $1.0$ to $1.8$ points over the GIN encoder that has no such option. How
much structure to use is decided during training rather than assumed.

\begin{table*}[t]
\caption{Module validation (macro-F1 \%, 20 seeds). $^{*}$ marks a variant significantly below the default at $p<0.05$ under a one-sided paired $t$-test over shared seeds.}
\label{tab:module-validation}
\centering
\resizebox{0.9\textwidth}{!}{
\begin{tabular}{c|cccc|cccc}
\toprule
Dataset & \multicolumn{4}{c|}{TwiBot-20} & \multicolumn{4}{c}{MGTAB} \\
\midrule
Setting & $\alpha=1.0$ & $\alpha=0.5$ & $\alpha=0.1$ & $\alpha=0.05$ & $\alpha=1.0$ & $\alpha=0.5$ & $\alpha=0.1$ & $\alpha=0.05$ \\
\midrule
\multicolumn{9}{c}{\emph{Client update-rate policy}} \\
\midrule
DDQN (default) & 85.3$\pm$0.7 & 84.6$\pm$1.3 & 84.0$\pm$1.7 & 83.3$\pm$1.9 & 85.8$\pm$1.2 & 84.3$\pm$1.7 & 80.3$\pm$2.8 & 77.4$\pm$3.3 \\
Softmax policy & 85.4$\pm$0.7 & 84.6$\pm$1.7 & 83.9$\pm$1.7 & 83.3$\pm$1.7 & 85.9$\pm$1.2 & 84.3$\pm$1.7 & 80.7$\pm$2.3 & 77.9$\pm$2.8 \\
Validation weighting & 85.5$\pm$0.7 & 83.9$\pm$2.2$^{*}$ & 83.3$\pm$1.6$^{*}$ & 82.4$\pm$2.4$^{*}$ & 85.2$\pm$1.1 & 84.5$\pm$1.3 & 79.8$\pm$4.2 & 76.8$\pm$2.7 \\
Random schedule & 85.4$\pm$0.7 & 84.4$\pm$1.5 & 83.9$\pm$1.8 & 83.1$\pm$2.0 & 85.6$\pm$1.0 & 84.5$\pm$1.4 & 79.8$\pm$3.3 & 78.1$\pm$1.9 \\
Fixed rate & 85.3$\pm$1.0 & 84.6$\pm$1.5 & 83.7$\pm$1.9$^{*}$ & 82.7$\pm$2.1$^{*}$ & 85.3$\pm$1.1 & 84.5$\pm$1.6 & 80.9$\pm$2.5 & 77.2$\pm$3.6 \\
\midrule
\multicolumn{9}{c}{\emph{Heterogeneous personalized heads}} \\
\midrule
Uniform heads (default) & 85.3$\pm$1.0 & 84.6$\pm$1.2 & 83.9$\pm$1.6 & 83.3$\pm$1.8 & 85.6$\pm$1.0 & 84.2$\pm$1.9 & 80.4$\pm$2.9 & 78.3$\pm$2.6 \\
Per-client head widths & 85.5$\pm$0.8 & 84.6$\pm$1.6 & 83.8$\pm$1.8 & 82.6$\pm$2.2 & 85.7$\pm$0.9 & 84.4$\pm$2.2 & 80.7$\pm$3.5 & 78.8$\pm$2.2 \\
\bottomrule
\end{tabular}}
\end{table*}

The upper block of Table~\ref{tab:module-validation} asks whether the controller needs
to be learned. Holding everything else fixed, we replace the DDQN by a softmax policy
over the same state, by weighting each client by its local validation accuracy, by a
random schedule, and by a fixed rate. The learned policy is not reliably better than the
simplest alternatives: across the sixteen policy--cell pairs formed by the softmax policy
and the random schedule, none is separable from the DDQN, all stay within $0.8$ macro-F1
points of it, and ten match or exceed it. The two that do lose sit at the extremes---a
fixed rate in two cells and validation weighting in three, all on TwiBot-20 and all under
skew, where a client's local validation split is least informative. Since a random
schedule ties while a fixed one loses, what the evidence supports is that the update rate
should \emph{vary} from round to round, not that it should be learned from the
federation's state.

The lower block asks what the personalization claim of Section~\ref{fedrio:method}
actually licenses. Giving each client a head of a different width, which the aggregation
permits because heads are never aggregated, leaves the eight cells within $-0.7$ to
$+0.5$ points with no significant difference. Width heterogeneity is therefore supported
at no measurable cost rather than beneficial, which is the narrow reading we adopt:
clients may differ in their heads and in their per-node message-passing modes, but not in
the shape of what they aggregate.

\vspace*{-3pt}
\subsection{Cross-Dataset Federation (Q6)}
\label{exp:crossplatform}

\begin{table*}[t]
\caption{Federations that span two datasets (macro-F1 \%). All methods see the $13$ account features the two datasets share. Left: trained on one dataset and evaluated once on the other, at $\alpha=0.5$ and $20$ seeds. Right: clients drawn from both datasets train together and are scored on the pooled test split, $10$ seeds. The first row is a constant predictor labelling every account with the majority class of the set it is scored on. Best per column in bold.}
\label{tab:crossplatform}
\centering
\resizebox{0.85\textwidth}{!}{
\begin{tabular}{c|cc|cc|cccc}
\toprule
& \multicolumn{4}{c|}{Held-out dataset (train $\to$ test)} & \multicolumn{4}{c}{Joint federation, pooled test split} \\
& \multicolumn{2}{c|}{TwiBot-20 $\to$ MGTAB} & \multicolumn{2}{c|}{MGTAB $\to$ TwiBot-20} & & & & \\
Method & source & held-out & source & held-out & $\alpha=1.0$ & $\alpha=0.5$ & $\alpha=0.1$ & $\alpha=0.05$ \\
\midrule
Majority class & 35.8 & 42.2 & 42.2 & 35.8 & 36.6 & 36.6 & 36.6 & 36.6 \\
\midrule
FedAvg & 77.8$\pm$1.0 & 51.1$\pm$6.9 & 52.1$\pm$14.2 & 31.8$\pm$2.8 & 79.5$\pm$0.9 & 73.5$\pm$13.4 & 50.7$\pm$15.8 & 50.4$\pm$15.4 \\
FedProx & 77.7$\pm$1.0 & 50.3$\pm$6.4 & 52.0$\pm$14.2 & 31.9$\pm$3.1 & 79.6$\pm$0.7 & 73.6$\pm$13.5 & 50.3$\pm$15.5 & 50.7$\pm$15.4 \\
FedDF & 78.1$\pm$1.1 & 49.3$\pm$6.8 & 51.7$\pm$14.1 & 31.5$\pm$2.0 & 79.4$\pm$0.9 & 72.0$\pm$13.6 & 53.2$\pm$14.1 & 50.9$\pm$15.3 \\
FedEnsemble & \textbf{78.2$\pm$0.9} & 45.3$\pm$7.6 & \textbf{64.2$\pm$15.6} & \textbf{36.2$\pm$6.8} & 79.4$\pm$0.6 & \textbf{76.7$\pm$5.2} & 66.7$\pm$11.7 & 60.0$\pm$14.7 \\
FedDistill & 78.0$\pm$1.0 & 49.5$\pm$7.1 & 51.1$\pm$13.8 & 31.5$\pm$1.9 & 79.6$\pm$0.6 & 71.9$\pm$13.5 & 49.1$\pm$14.8 & 51.7$\pm$14.9 \\
FedGen & 77.9$\pm$1.0 & 50.8$\pm$6.8 & 52.2$\pm$14.2 & 32.0$\pm$3.4 & 79.4$\pm$0.9 & 73.4$\pm$13.5 & 51.5$\pm$15.9 & 50.5$\pm$15.4 \\
FedFTG & 78.1$\pm$1.0 & 49.1$\pm$6.0 & 56.1$\pm$15.4 & 31.8$\pm$2.7 & 79.4$\pm$0.8 & 71.4$\pm$13.7 & 53.3$\pm$16.2 & 51.9$\pm$15.1 \\
GCFL+ & 73.5$\pm$3.0 & 49.4$\pm$5.7 & 50.0$\pm$5.5 & 34.1$\pm$4.0 & 75.7$\pm$3.1 & 72.4$\pm$4.8 & 68.5$\pm$4.1 & 70.0$\pm$4.3 \\
FED-PUB & 73.9$\pm$3.2 & 50.1$\pm$7.3 & 50.5$\pm$7.8 & 32.3$\pm$1.8 & 72.8$\pm$7.7 & 63.7$\pm$11.7 & 59.4$\pm$10.5 & 56.8$\pm$16.4 \\
\FedACK & 77.9$\pm$0.9 & \textbf{52.2$\pm$6.8} & 52.3$\pm$12.8 & 31.3$\pm$1.4 & \textbf{80.1$\pm$0.7} & 67.3$\pm$17.4 & 52.0$\pm$14.1 & 53.0$\pm$16.6 \\
\midrule
\FedRio & 75.8$\pm$2.4 & 51.9$\pm$7.3 & 60.4$\pm$8.0 & 35.8$\pm$2.8 & 77.2$\pm$2.7 & 74.1$\pm$4.7 & \textbf{73.2$\pm$2.3} & \textbf{73.6$\pm$3.3} \\
\bottomrule
\end{tabular}}
\vspace{-3pt}
\end{table*}

The Dirichlet protocol varies label distributions while drawing every client from a single
user population. Table~\ref{tab:crossplatform} replaces that assumption with two settings
whose clients come from two independently collected datasets. The datasets differ in
language coverage, relation schema, label prior, graph density, and collection period.
Neither setting is expressible in the feature space used so far, since the two benchmarks embed text with different sentence
encoders and record different relation types, so both restrict every method to the $13$
account properties the benchmarks record in common. That view is far poorer than the $794$
and $788$ features of Section~\ref{exp:setup}, so scores here are lower throughout and
only comparisons within the table are meaningful.

The left half trains a federation on one dataset and evaluates its global model once on
the other, which never participates. Transfer is weak and asymmetric. Out of TwiBot-20,
every method attains between $45.3$ and $52.2$ macro-F1 against the $42.2$ of a
majority-class predictor on MGTAB, and \FedRio's $51.9$ trails \FedACK\ by less than any
paired test resolves. The reverse direction transfers nothing at all: the field spans
$31.3$ to $36.2$ against a majority baseline of $35.8$, so even its best entry,
FedEnsemble's $36.2$, only exceeds by $0.4$ what a predictor that ignores its input
already achieves.

\begin{table*}[t]
\caption{Adapting to a held-out dataset (macro-F1 \%, $20$ seeds). The source-trained encoder is frozen and a logistic-regression head refitted on $k$ labeled target accounts; $k=0$ is the zero-shot read-out. Supports are class-balanced, nested and shared across methods, and scoring excludes every support account, so the $k=0$ column differs marginally from the held-out column of Table~\ref{tab:crossplatform}. ``dim'' is the effective number of directions the representation retains on the target, out of $64$. Best per column in bold.}
\label{tab:kshot}
\centering
\resizebox{0.9\textwidth}{!}{
\begin{tabular}{c|ccccc|ccccc}
\toprule
& \multicolumn{5}{c|}{TwiBot-20 $\to$ MGTAB} & \multicolumn{5}{c}{MGTAB $\to$ TwiBot-20} \\
Method & $k{=}0$ & $k{=}10$ & $k{=}50$ & $k{=}100$ & dim & $k{=}0$ & $k{=}10$ & $k{=}50$ & $k{=}100$ & dim \\
\midrule
FedAvg & 51.1$\pm$6.9 & 57.3$\pm$5.9 & 67.1$\pm$3.1 & 70.6$\pm$2.6 & 30.0 & 31.8$\pm$2.8 & 65.6$\pm$7.4 & 70.9$\pm$3.8 & 73.2$\pm$2.0 & 29.4 \\
FedProx & 50.2$\pm$6.4 & 57.1$\pm$5.5 & 66.7$\pm$3.1 & 70.6$\pm$2.7 & 29.6 & 31.9$\pm$3.1 & 65.6$\pm$7.1 & 70.9$\pm$3.8 & 73.2$\pm$2.1 & 29.6 \\
FedDF & 49.2$\pm$6.9 & 56.1$\pm$5.8 & 66.2$\pm$3.6 & 70.1$\pm$2.6 & 27.5 & 31.5$\pm$2.0 & \textbf{66.1$\pm$7.7} & 70.8$\pm$3.7 & 73.2$\pm$2.0 & 28.7 \\
FedEnsemble & 45.2$\pm$7.6 & 56.3$\pm$5.4 & 66.7$\pm$2.9 & 70.7$\pm$2.4 & 30.6 & \textbf{36.2$\pm$6.8} & 65.9$\pm$7.1 & 71.0$\pm$3.8 & 73.3$\pm$2.0 & 29.1 \\
FedDistill & 49.4$\pm$7.1 & 57.3$\pm$5.2 & 66.9$\pm$3.3 & 70.5$\pm$2.9 & 30.8 & 31.5$\pm$2.0 & 65.5$\pm$7.0 & 70.8$\pm$3.7 & 73.1$\pm$1.9 & 30.3 \\
FedGen & 50.8$\pm$6.8 & 57.1$\pm$5.8 & 66.7$\pm$3.2 & 70.3$\pm$2.6 & 30.1 & 32.0$\pm$3.4 & 65.5$\pm$7.5 & \textbf{71.1$\pm$3.9} & \textbf{73.4$\pm$2.1} & 29.5 \\
FedFTG & 49.0$\pm$6.0 & 56.7$\pm$6.5 & 66.6$\pm$3.4 & 70.3$\pm$2.8 & 27.0 & 31.8$\pm$2.7 & 64.8$\pm$8.5 & 70.9$\pm$3.7 & 72.9$\pm$2.5 & 26.9 \\
GCFL+ & 49.3$\pm$5.7 & 57.0$\pm$7.1 & 67.4$\pm$4.0 & 71.6$\pm$2.0 & 34.8 & 34.1$\pm$4.0 & 58.2$\pm$7.7 & 68.7$\pm$4.7 & 71.9$\pm$2.5 & 25.6 \\
FED-PUB & 50.1$\pm$7.4 & 57.2$\pm$8.2 & \textbf{67.9$\pm$3.9} & \textbf{71.8$\pm$1.9} & 31.6 & 32.3$\pm$1.8 & 63.7$\pm$7.5 & 70.7$\pm$3.9 & 72.9$\pm$2.2 & 32.2 \\
\FedACK & \textbf{52.2$\pm$6.8} & \textbf{59.4$\pm$4.3} & 66.7$\pm$3.0 & 70.3$\pm$2.7 & 22.2 & 31.3$\pm$1.4 & 64.3$\pm$7.9 & 70.5$\pm$3.9 & 73.1$\pm$2.0 & 28.2 \\
\midrule
\FedRio & 51.8$\pm$7.4 & 56.9$\pm$7.0 & 67.0$\pm$2.6 & 70.5$\pm$2.3 & 33.8 & 35.7$\pm$2.8 & 62.9$\pm$9.4 & 70.4$\pm$3.7 & 72.9$\pm$1.5 & 33.2 \\
\bottomrule
\end{tabular}}
\end{table*}

Zero-shot transfer, however, is not the only deployment scenario for a newly collected
dataset: a small labeled support set may be available, and the question is which federated
model provides the best starting point. Table~\ref{tab:kshot} answers this. We take the model each method
trained on the source dataset, freeze its encoder, and refit only the logistic-regression
head on $k$ labeled accounts of the target, for $k$ of ten, fifty and a hundred. The
labeled accounts are the same for every method, and scoring uses only target accounts
that were never labeled, so what the comparison measures is the transferred
representation rather than a fine-tuning recipe. A hundred labels take every method well
past the zero-shot figures above, to between $70.1$ and $73.4$, and about one point then
separates the field: where the source is TwiBot-20, GCFL+ and FED-PUB finish $1.1$ and
$1.3$ points above \FedRio\ (both $p<0.01$), while in the opposite direction no baseline
differs significantly from it. The labels therefore matter far more than which method
produced the encoder, and where that choice still matters it does not favor ours.

One quantity does distinguish the methods. The last column of Table~\ref{tab:kshot},
marked ``dim'', reports the effective rank~\cite{roy2007effective} of each representation
on the dataset it was not trained on: the exponential of the entropy of its normalized
singular-value spectrum, which is $64$ when all directions carry equal weight and
approaches $1$ as the representation collapses onto a few. \FedRio\ retains $33.8$ and
$33.2$ in the two directions, significantly more than nine of the ten baselines in the
first and eight of ten in the second, and is never significantly below any of them;
\FedACK\ retains $22.2$, the lowest in the table, while matching \FedRio\ zero-shot out of
TwiBot-20 and falling $4.5$ points behind it out of MGTAB. Resisting collapse is what one
would want from a detector meant to be redeployed, but we state its value precisely: it
does not imply a higher score, since of the two methods above \FedRio\ at a hundred labels
one retains a wider representation and the other a narrower one.

The right half of Table~\ref{tab:crossplatform} instead has clients drawn from both datasets
train together and scores the model on the pooled test split. There the pattern of Q1
reappears in a stronger form. With labels spread evenly the dataset mixture is not the
binding constraint and \FedRio\ trails the simple baselines ($77.2$ against $80.1$); as
the partition sharpens the baselines collapse---FedAvg falls from $79.5$ to $50.4$---while
\FedRio\ holds, finishing $4.7$ and $3.6$ points above the strongest baseline at
$\alpha=0.1$ and $\alpha=0.05$. The per-dataset read-out, which scores the joint model
separately on clients drawn from each dataset, locates the remaining weakness: at $\alpha=0.05$
\FedRio\ attains $73.3$ on the TwiBot-20 clients against $59.5$ on the MGTAB clients, so
the joint model is carried by the larger dataset and clients from the smaller dataset
remain under-served.

\vspace*{-3pt}
\subsection{Comparison with Centralized Detectors (Q7)}
\label{sec:discussion}

Our baselines are federated methods, which are the methodologically comparable target,
but a reader is entitled to ask what federation costs against a detector that may hold
the whole graph. Table~\ref{tab:contextual_comparison} places \FedRio\ next to
published centralized results on TwiBot-20. The comparison is contextual: those numbers
come from the original papers, are produced with full data access, and are not rerun
here.

\begin{table}[t]
  \centering
  \caption{Contextual comparison between centralized social bot detectors (published
  results on TwiBot-20 with full data access) and \FedRio\ (federated, each client
  holding only its own accounts and the edges among them). Centralized results are
  quoted from the original papers; ``$\dagger$'' marks numbers quoted from
  LMBot~\cite{cai2024lmbot}.}
  \label{tab:contextual_comparison}
  \resizebox{0.48\textwidth}{!}{
  \begin{tabular}{l|c|cc|l}
  \toprule
  Method & Setting & Acc.(\%) & F1(\%) & Data access \\
  \midrule
  \multicolumn{5}{l}{\textit{Centralized, complete graph}} \\
  \midrule
  SGBot$^\dagger$ & Central. & 79.58 & 83.51 & Full graph + tweets \\
  BotRGCN$^\dagger$ & Central. & 84.42 & 86.90 & Full graph + metadata \\
  RGT$^\dagger$ & Central. & 84.70 & 87.19 & Full graph + metadata \\
  SimpleHGN$^\dagger$ & Central. & 84.65 & 87.13 & Full heterog. graph \\
  LMBot$^\dagger$ & Central. & 85.63 & 87.61 & Full graph $\to$ LM distill. \\
  SeBot~\cite{yang2024sebot} & Central. & 87.24 & 88.74 & Full graph + struct. entropy \\
  BotDGT~\cite{he2024dynamicity} & Central. & \textbf{87.25} & \textbf{88.87} & Full dynamic graph \\
  \midrule
  \multicolumn{5}{l}{\textit{Federated, 10 clients, no cross-client edges}} \\
  \midrule
\FedACK\ ($\alpha=1$) & Feder. & 84.52 & 84.24 & Local subgraph only \\
FedDistill ($\alpha=1$) & Feder. & 85.77 & 85.44 & Local subgraph only \\
  \FedRio\ ($\alpha=1$) & Feder. & 85.66 & 85.29 & Local subgraph only \\
  \FedRio\ ($\alpha=0.05$) & Feder. & 83.90 & 83.26 & Local subgraph only \\
  \bottomrule
  \end{tabular}
  }
\end{table}

Three observations follow. First, the cost of federation on this benchmark is smaller
than the spread among centralized methods themselves: \FedRio\ at $\alpha=1$ reaches
$85.66\%$ accuracy against $87.25\%$ for the strongest centralized detector, a gap of
$1.6$ points, while the centralized methods span $7.7$ points among themselves. Second,
the comparison is not like for like and we do not present it as one---the centralized
detectors see every edge, including the cross-client edges that federation removes by
construction, and several of them additionally use tweet text or temporal structure
that our common feature view does not carry. Third, the interesting quantity is not the
gap at $\alpha=1$ but its behavior under skew: \FedRio\ retains $83.90\%$ at
$\alpha=0.05$, so the distance to a centralized detector grows by only $1.8$ points
when the federation moves from near-uniform to extremely skewed clients, whereas the
federated baselines lose between $5.7$ and $16.4$ points over the same range.

Running the centralized detectors inside our federation would not sharpen this
comparison. Their contribution derives from inputs the federated setting removes:
BotRGCN and RGT operate on the multi-relational graph, SeBot on structural entropy over
the whole graph, and BotDGT on temporal dynamics. A client in our partitions holds
between $6.6\%$ and $25.7\%$ (Section~\ref{setup:graph}), so each
of these methods would be handed a fraction of the structure it was designed to exploit,
and the outcome would report the limitations of our setting as though they were
limitations of those methods.
\vspace{-2pt}
\subsection*{Limitations and Future Work}
Three limitations follow from our results rather than from the design. The first concerns
which component earns the gain, and the answer depends on the partition: the two
mechanisms we propose separate from seed noise under moderate skew, whereas on the
sharpest MGTAB partitions it is the cooperative encoder that carries the margin. No single
component therefore explains \FedRio\ across the whole range of skew we test. 
The same caution applies to the client controller: replacing its learned policy with a random schedule changes nothing a paired test detects, while a fixed rate does lose, so what our evidence supports is that the update rate should vary from round to round and not that it should be learned from the federation's state.
The second is that a model trained on one dataset transfers poorly to a dataset that did not participate in
the federation. This holds for every method we tested, and in one of the two directions no
method beats a predictor that ignores its input. Table~\ref{tab:kshot} shows where the
remedy lies: a hundred labeled accounts from the target dataset close most of the gap, and
once they are available the choice of federated method is worth about one point. The third
is one of scope. Our graphs are the relations these two benchmarks happen to record, and
both were collected from Twitter, so our federations span independently collected datasets
rather than different platforms. We found no public non-Twitter benchmark that combines
a recorded social graph, account-level bot labels, and raw content from which a shared
feature space could be built. Whether the ranking reported here holds across platforms,
or on a larger and more recent benchmark such as TwiBot-22~\cite{feng2022twibot}, remains
to be tested.

\section{Conclusion}
\label{fedrio:conclusion}
This paper presents \FedRio, a personalized federated framework for social bot detection
across platforms that cannot pool their data. Combining a cooperative encoder that lets
each account select how it propagates information through the fragment of the graph its
holder observes, data-free adversarial distillation with contrastive alignment,
neuron-level server aggregation, and a reinforcement-controlled client update rate,
\FedRio\ improves the detection quality a federation obtains under label skew: over twenty
seeds with validation-selected checkpoints and paired tests, it leads the strongest
federated baseline in six of eight dataset--skew configurations by $1.4$ to $7.4$ macro-F1
points, widens that margin as labels spread more unevenly across clients on both
benchmarks, and leads a federation spanning two datasets once those labels are uneven.
\FedRio\ is therefore the stronger choice when a federation is judged by the model it
produces jointly, whereas a participant judged only on its own accounts under extreme
skew is better served by a personalization-first method.
Which of these components is worth its cost turns out to depend on how unevenly the labels
are spread, and establishing that dependence for federated detectors more generally is the
direction we intend to pursue.